\documentclass[sigconf,nonacm]{acmart}

\usepackage{tikz}
\usepackage{booktabs}
\usepackage{algorithm}
\usepackage{algpseudocode}
\usepackage{xspace}
\usepackage{tabularx}
\usepackage{colortbl}

\newcommand{\router}{PISA\xspace}
\newcommand{\routerM}{PISA-E\xspace}

\usepackage{dsfont}
\usepackage{multirow}

\usetikzlibrary{arrows.meta,calc,backgrounds}

\AtBeginDocument{%
  }

\setcopyright{none}
\acmYear{2026}
\acmDOI{XXXXXXX.XXXXXXX}
\acmConference[WWW '27]
  {The ACM Web Conference 2027}
  {May 10--14, 2027}
  {Dublin, Ireland}

\acmSubmissionID{4320}

\begin{document}

\title{Ensemble Routing for Prompt-Adaptive Cost-Aware LLM Inference}

\author{
  Saul Urso$^{1,3}$,
  Emanuele Carlini$^{1}$,
  Patrizio Dazzi$^{1,3}$,
  Matteo Mordacchini$^{2}$,
}

\affiliation{%
  \institution{%
    $^{1}$Institute of Information Science and Technologies, CNR, Italy\\
    $^{2}$Institute of Informatics and Telematics, CNR, Italy\\
    $^{3}$University of Pisa, Department of Computer Science, Italy
  }
  \country{}
}

\email{%
  saul.urso@phd.unipi.it,
  emanuele.carlini@isti.cnr.it,
  patrizio.dazzi@unipi.it,
  matteo.mordacchini@iit.cnr.it
}

\renewcommand{\shortauthors}{Urso et al.}




\begin{abstract}

Large language models (LLMs) increasingly support Web-based services that must balance response quality and inference cost across heterogeneous requests. Conventional routing selects one model per prompt, while fixed ensembles cannot adapt their composition and cost to individual requests. We introduce \router, a prompt-adaptive, cost-aware single-model router, and \routerM, its extension to ensemble routing. At the core of both methods, \router combines soft cluster memberships with model-specific predictors to estimate individual model capabilities. Building on these estimates, \routerM jointly selects ensemble composition and cardinality under an explicit cost--quality preference that can be adjusted at inference time without retraining. A tractable ensemble-success surrogate and efficient candidate selection enable this extension without ensemble-specific training. Across four LLM routing benchmarks under in-distribution (ID) and out-of-distribution (OOD) conditions, \router achieves the highest macro-average cost--performance hypervolume among the evaluated single-model routers in both settings. Meanwhile, \routerM achieves the highest macro-average task performance among the evaluated ensemble approaches when prioritizing response quality, surpassing the strongest competitors by 2.09 and 2.87 percentage points under ID and OOD conditions, respectively. At selected lower-cost operating points, it also exceeds the strongest ensemble competitors' macro-average task performance while reducing average inference cost by approximately 36\% and 25\%, respectively. These results demonstrate the effectiveness of prompt-adaptive capability estimation for both single-model routing and selective ensembling.

\end{abstract}

\maketitle


\section{Introduction}
\label{sec:introduction}

Large language models (LLMs) are increasingly deployed in Web-based services, supporting applications such as search and information discovery, recommendation, and intelligent agents \cite{hui2026realtime,behdin2025scaling}. Their adoption also extends to publicly accessible LLM services, creating new challenges for efficient and scalable inference \cite{liu2026exploring}. These services must handle heterogeneous user requests while balancing response quality and inference cost. When multiple LLMs with different capabilities and pricing structures are available, invoking expensive models indiscriminately wastes resources, whereas consistently selecting inexpensive models can compromise response quality. Moreover, different users and requests may have different cost preferences, making it desirable to adapt inference strategies to both prompt characteristics and inference-cost preferences~\cite{varangot2026doing}.

LLM routing addresses this challenge by selecting models according to prompt characteristics, expected response quality, and inference cost~\cite{moslem2026dynamic}. Recent approaches extend routing to budget-adaptive selection~\cite{zu2026BARouter}, reasoning-strategy selection~\cite{pan2026Adaptive}, and additional deployment constraints~\cite{li2026Fair}. However, selecting a single model cannot exploit the complementary capabilities of multiple LLMs. Ensembling offers an alternative by combining multiple model responses~\cite{chen2026Harnessing}, but introduces additional inference costs. Existing ensemble approaches explore adaptive model composition, diversity-aware selection, and budget-constrained optimization~\cite{huang2024Ensemble,cohen2026DFPE,tekin2024LLMTOPLA}.
Nevertheless, existing approaches typically address only parts of the problem: some adapt ensemble composition without explicitly optimizing inference cost, while others rely on classification-specific aggregation or sequential execution. Moreover, ensemble cardinality is generally fixed during training or configuration, limiting the ability to adjust the number of invoked models to individual prompts and changing cost preferences at inference time. This motivates \emph{ensemble routing}, where both the composition and cardinality of the invoked model subset are determined dynamically at inference time according to prompt characteristics and an explicit cost--quality preference.

We introduce \emph{\router}, a prompt-adaptive, cost-aware single-model router that estimates the capabilities of individual LLMs, and \emph{\routerM}, its extension to prompt-adaptive ensemble routing.
\routerM reuses the individual capability estimates to construct a tractable surrogate for ensemble success, eliminating the need to train separate quality estimators for different model combinations. By combining these estimates with inference cost, \routerM selects prompt-specific model subsets, jointly determining which models to invoke and how many. Crucially, the cost--quality preference can be adjusted at inference time without additional hyperparameter optimization or retraining, allowing the same framework to support different operating points. The selected responses are combined through capability-weighted semantic similarity to produce the final ensemble response.

We evaluate \router and \routerM on four LLM routing benchmarks under in-distribution (ID) and out-of-distribution (OOD) conditions. \router achieves the highest macro-average hypervolume among the evaluated single-model routers in both settings, ranking first in six of the eight benchmark–setting combinations. \routerM achieves the highest macro-average task performance among the evaluated ensemble approaches in both settings, improving over the strongest competitors by 2.09 percentage points under ID and 2.87 under OOD conditions when prioritizing response quality. At selected lower-cost operating points, it also exceeds the strongest competitors’ macro-average task performance while reducing average inference cost by approximately 36\% and 25\%, respectively. Finally, comparisons with fixed-cardinality variants demonstrate the benefits of adapting ensemble size to individual prompts.


Our main contributions are:
\begin{itemize}
    \item We develop \textbf{\router}, a prompt-adaptive, cost-aware single-model router that estimates individual model capabilities through soft cluster memberships and model-specific predictors, providing the foundation for ensemble routing.
    \item We propose \textbf{\routerM}, a framework for prompt-adaptive, cost-aware ensemble routing that uses ensemble-level utility estimates derived from individual model capabilities to jointly determine ensemble composition and cardinality at inference time, according to prompt characteristics and inference-cost preferences.
    \item We evaluate both methods across four benchmarks under ID and OOD conditions, demonstrating the highest macro-average single-model routing HV and ensemble task performance among the evaluated approaches in both settings. We also demonstrate operating points with higher task performance at lower cost than the strongest ensemble competitors and quantify the benefits of prompt-dependent ranking and adaptive ensemble cardinality.
\end{itemize}
\section{Problem Formulation}
\label{sec:problem-formulation}

Let $\mathcal{M}=\{1,\ldots,M\}$ denote a fixed pool of candidate LLMs. Given a prompt $q$, model routing selects one or more models from $\mathcal{M}$ while trading off response quality and inference cost.

For each model $m\in\mathcal{M}$, let $X_m(q)\in\{0,1\}$ indicate whether the model produces a correct answer, with prompt-dependent probability
\begin{equation}
    p_m(q)
    =
    \Pr\!\left(X_m(q)=1\mid q\right).
    \label{eq:true-model-probability}
\end{equation}
Since $p_m(q)$ is unknown at inference time, a router operates on an estimate $\hat p_m(q)$. Let $\bar c_m(q)\in[0,1]$ denote the normalized cost of querying model $m$ on prompt $q$.

\subsection{Cost-Aware Single-Model Routing}
\label{sec:single-routing}

Standard model routing selects exactly one model for each prompt. We formulate the cost-aware routing decision as
\begin{equation}
    \hat m(q)
    =
    \operatorname*{arg\,max}_{m\in\mathcal{M}}
    J_m(q;\alpha),
    \label{eq:single-model-decision}
\end{equation}
where
\begin{equation}
    J_m(q;\alpha)
    =
    \alpha \hat p_m(q)
    +
    (1-\alpha)\left(1-\bar c_m(q)\right),
    \qquad \alpha\in[0,1].
    \label{eq:single-model-utility}
\end{equation}
Here, $\alpha$ controls the trade-off between predicted response quality and inference cost.

\subsection{Cost-Aware Ensemble Routing}
\label{sec:multi-routing-formulation}

We generalize standard routing by allowing the router to select any non-empty subset
$
    E\subseteq\mathcal{M},
     E\neq\emptyset
$. All models in $E$ are queried, and their outputs are combined by a response resolver $\mathcal{A}$, which may implement any suitable aggregation strategy. If $Y_m(q)$ denotes the output of model $m$, the resulting ensemble output is

\begin{equation}
    Y_E(q)
    =
    \mathcal{A}\!\left(\{Y_m(q):m\in E\}\right).
\end{equation}
Let
\begin{equation}
    X_E(q)
    =
    \mathds{1}\!\left[Y_E(q)\text{ is correct}\right],
\end{equation}
and define the corresponding probability of correctness as
\begin{equation}
    p_E(q)
    =
    \Pr\!\left(X_E(q)=1\mid q\right).
    \label{eq:true-ensemble-probability}
\end{equation}

As in the single-model setting, $p_E(q)$ is unavailable at inference time; we denote its router-provided estimate by $\hat p_E(q)$. Let $\bar C(E,q)\in[0,1]$ denote the normalized inference cost associated with querying ensemble $E$. Cost-aware ensemble routing then solves
\begin{equation}
    \hat E(q)
    =
    \operatorname*{arg\,max}_{\emptyset\neq E\subseteq\mathcal{M}}
    J_E(q;\alpha),
    \label{eq:multi-model-decision}
\end{equation}
with
\begin{equation}
    J_E(q;\alpha)
    =
    \alpha \hat p_E(q)
    +
    (1-\alpha)\left(1-\bar C(E,q)\right).
    \label{eq:ensemble-utility}
\end{equation}

Single-model routing is recovered by restricting the feasible ensembles to singletons, $|E|=1$. Thus, the two settings differ only in their action spaces: standard routing selects an element of $\mathcal{M}$, whereas ensemble routing selects an element of $2^{\mathcal{M}}\setminus\{\emptyset\}$.

Solving this problem requires estimating the expected quality and inference cost of candidate model subsets. Section~\ref{sec:methodology} describes how we instantiate these quantities and perform cost-aware model and ensemble selection.

\begin{figure*}[t]
    \centering
    \resizebox{\linewidth}{!}{%
    \input{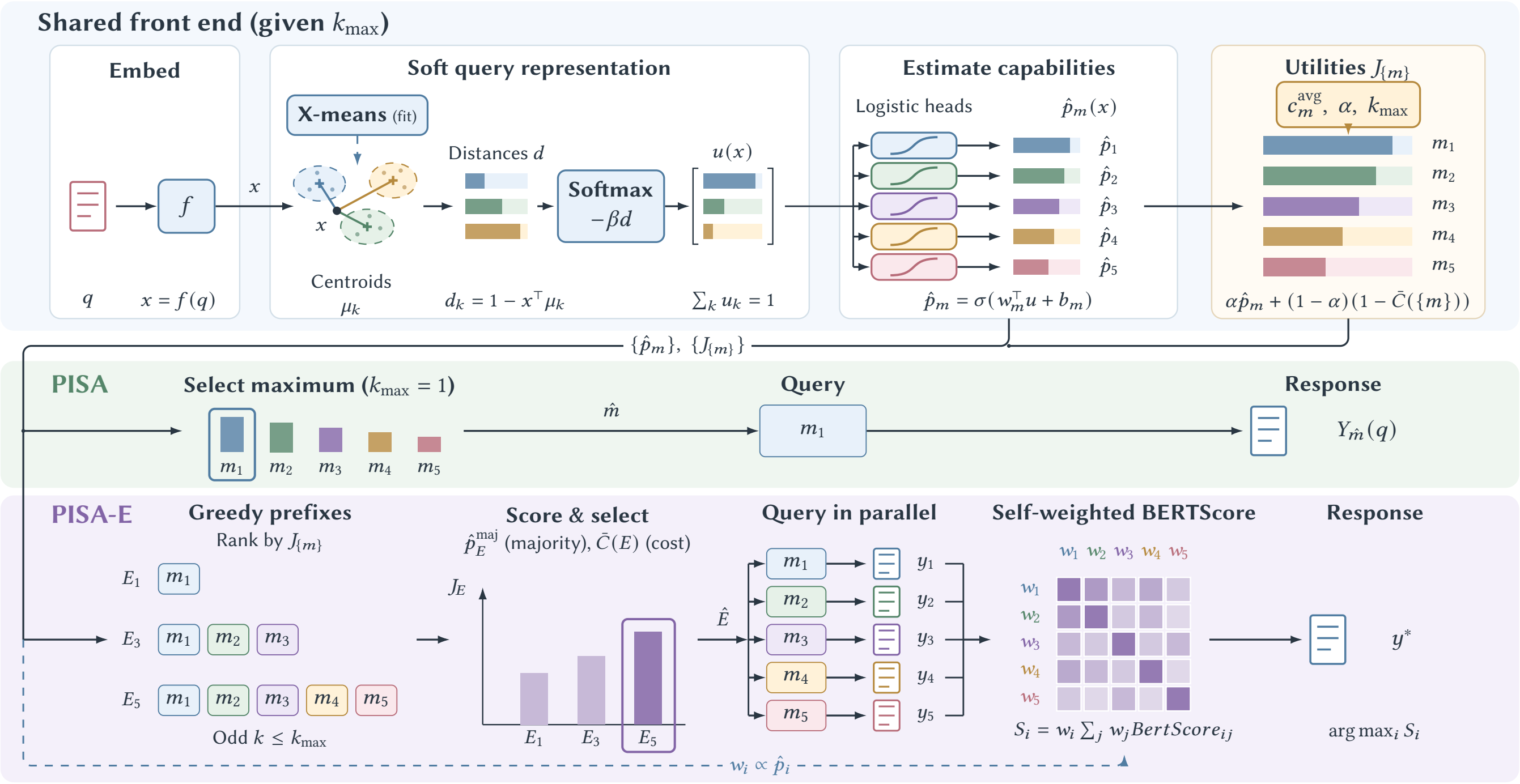}%
    }
    \Description{}
    \caption{\textbf{Inference-time overview of \router and \routerM.} \router maps a prompt embedding to soft memberships over X-means centroids, estimates each model's success probability, and selects the model with highest cost-aware utility. \routerM reuses the capability estimates to rank models by utility using its ensemble cost normalization, and evaluates odd-cardinality ranking prefixes using a strict-majority success surrogate and normalized ensemble cost. In the illustrated five-model example, the candidates are $E_1$, $E_3$, and $E_5$. The selected ensemble is queried in parallel, and self-weighted BERTScore selects the final response using the estimated model capabilities.}

    \label{fig:routing-overview}
\end{figure*}

\section{Methodology}
\label{sec:methodology}

Our approach builds on a shared capability-estimation framework to support both single-model and ensemble routing. \router estimates the prompt-dependent capabilities of individual LLMs and uses them for cost-aware single-model selection. \routerM extends this framework to ensemble routing by combining the same capability estimates into an ensemble-success surrogate, allowing model subsets of different compositions and cardinalities to be evaluated under the cost--quality objective introduced in Section~\ref{sec:problem-formulation}. Figure~\ref{fig:routing-overview} illustrates the resulting inference pipeline.

\subsection{\router: Model Routing via Soft Assignments}
\label{sec:router}

The first stage estimates how likely each candidate model is to produce a correct response to a given prompt. \router learns prompt-dependent capability estimates from historical model outcomes and semantic representations of the input queries. Assume a training set of $N$ prompts, where each prompt $q_i$ is represented by an embedding $x_i=f(q_i)\in\mathbb{R}^d$ and is associated with a correctness score $s_{im}\in[0,1]$ for each model $m\in\mathcal{M}$. At inference time, only the embedding $x=f(q)$ of the incoming prompt is available.

To capture similarities among prompts, we partition the fit-set embeddings using X-means, which determines both the number of clusters $K$ and their centroids. \router represents each query through soft memberships over the learned centroids, allowing its representation to incorporate information from multiple clusters. After $\ell_2$-normalizing embeddings and centroids, these memberships are computed from cosine distances as
\begin{equation}
    u_k(x)
    =
    \frac{\exp(-\beta d_k(x))}
         {\sum_{j=1}^{K}\exp(-\beta d_j(x))},
    \qquad
    d_k(x)=1-x^\top\mu_k,
    \label{eq:softmax-membership}
\end{equation}
where $\mu_k$ denotes the centroid of cluster $k$ and $\beta>0$ controls the concentration of the assignments. This representation allows capability estimation to exploit information from multiple neighboring regions instead of relying on a single hard assignment.

Using the resulting membership vector $u(x)$, \router learns an independent logistic predictor for each candidate model. Each predictor is trained using the soft memberships of the training prompts and their corresponding binary correctness labels, obtained by thresholding evaluation scores at $0.5$. For an incoming prompt, the estimated capability of model $m$ is
\begin{equation}
    \hat p_m(x)
    =
    \sigma\!\left(
        w_m^\top u(x) + b_m
    \right),
    \label{eq:router-capability}
\end{equation}
where $\sigma(\cdot)$ is the logistic sigmoid and $(w_m,b_m)$ are the parameters fitted for model $m$ using L2-regularized logistic regression. The regularization strength $\lambda_m$ is selected independently for each model through 5-fold cross-validation minimizing log loss over 10 logarithmically spaced values between $10^{-4}$ and $10^{4}$. The selected predictor is then refitted on the full training set.

To incorporate inference cost into the routing decision, we estimate each model's cost from its average observed training-set cost. Let $c_{im}$ denote the cost of invoking model $m$ on training prompt $q_i$, and define
\begin{equation}
    c_m^{\mathrm{avg}}
    =
    \frac{1}{N}\sum_{i=1}^{N}c_{im},
    \qquad
    \bar c_m
    =
    \frac{c_m^{\mathrm{avg}}}
         {\max_{m'\in\mathcal{M}}c_{m'}^{\mathrm{avg}}}.
    \label{eq:single-model-cost-normalization}
\end{equation}
The normalized cost $\bar c_m\in[0,1]$ provides the cost term in Eq.~\eqref{eq:single-model-utility}. For each incoming prompt and cost preference $\alpha$, \router selects the model maximizing the corresponding single-model utility in Eq.~\eqref{eq:single-model-decision}. The resulting capability estimates also provide the foundation for the ensemble-routing extension described next.

\subsection{\routerM: Prompt-Adaptive Cost-Aware LLM Ensembling}
\label{sec:routerM}

While \router selects a single model, \routerM extends the routing decision to model subsets, allowing both ensemble composition and cardinality to vary with the prompt and cost preference. Given the capability estimates $\{\hat p_m(x)\}_{m=1}^{M}$, \routerM considers ensembles $E\subseteq\mathcal{M}$ with $1\leq |E|\leq k_{\max}$, where $k_{\max}$ denotes the maximum ensemble size. Its central challenge is to evaluate the expected utility of candidate subsets before invoking their constituent models.

Directly estimating the performance of every possible ensemble would require modeling the behavior of numerous model combinations. Instead, \routerM constructs an ensemble-level utility estimate from the individual capabilities already provided by \router. Specifically, we use the probability that a strict majority of the selected models produce correct responses as a tractable surrogate for comparing candidate ensembles. This criterion captures how the predicted capabilities of the constituent models combine, without requiring ensemble-specific predictors or observing their generated responses.

For an ensemble $E$ of size $k$, we model the correctness of each constituent model through a Bernoulli variable $Z_m$ with success probability $\hat p_m(x)$. Assuming conditional independence of these variables given the prompt, the strict-majority success probability is
\begin{equation}
    \hat p_E^{\mathrm{maj}}(x)
    =
    \Pr\!\left(
        \sum_{m\in E} Z_m
        \geq
        \left\lfloor\frac{k}{2}\right\rfloor+1
    \right),
    \quad
    Z_m\sim\operatorname{Bernoulli}(\hat p_m(x)),
    \label{eq:predicted-majority-probability}
\end{equation}
with the $Z_m$ treated as independent. We compute this probability using the standard Poisson-binomial dynamic program. The strict-majority success probability serves as a surrogate for ensemble selection, since the outcome of the final response resolver also depends on relationships among the generated responses. Accordingly, $\hat{p}^{\mathrm{maj}}_E(x)$ provides the quality term in the ensemble-routing objective independently of the response resolution procedure.

To balance this surrogate with inference cost, \routerM estimates the cost of an ensemble by summing the average costs of its constituent models. Let the model costs defined in Eq.~\eqref{eq:single-model-cost-normalization} be ordered as $c_{(1)}^{\mathrm{avg}}\geq\cdots\geq c_{(M)}^{\mathrm{avg}}$. We normalize the ensemble cost as
\begin{equation}
    \bar C(E)
    =
    \frac{
        \sum_{m\in E}c_m^{\mathrm{avg}}
    }{
        \sum_{j=1}^{k_{\max}}c_{(j)}^{\mathrm{avg}}
    }.
    \label{eq:predicted-normalized-ensemble-cost}
\end{equation}
The denominator corresponds to the maximum cost of any subset containing at most $k_{\max}$ models and remains constant across candidate ensembles. Combining this normalized cost with the strict-majority surrogate yields the ensemble utility in Eq.~\eqref{eq:ensemble-utility}, allowing subsets of different cardinalities to be compared under the same cost--quality preference.

Since evaluating all feasible subsets would require a combinatorial search, \routerM uses a greedy candidate-construction procedure. For each prompt, the models are first ranked according to their utilities, using the same cost normalization as for all candidate ensembles:
\begin{equation}
    J_{\{m\}}(x;\alpha)
    = \alpha\hat p_m(x)
    +(1-\alpha)\bigl(1-\bar C(\{m\})\bigr).
    \label{eq:ensemble-ranking-utility}
\end{equation}
Here, $\bar C(\{m\})$ is obtained from Eq.~\eqref{eq:predicted-normalized-ensemble-cost} with $E=\{m\}$, so the cost of each model is normalized by the sum of the $k_{\max}$ largest average model costs. Candidate ensembles are then constructed as progressively larger prefixes of this ranking, considering odd cardinalities up to $k_{\max}$. Each candidate is evaluated using its strict-majority success probability and normalized inference cost, and the ensemble maximizing the resulting utility is selected. For $k_{\max}=1$, both the cost normalization and the selection rule reduce to those of \router.

Restricting the candidates to odd cardinalities follows directly from the strict-majority surrogate. For any even-cardinality ensemble, removing one constituent model produces an odd-cardinality subset whose strict-majority success probability is at least as high, since the required number of correct models decreases by one. As the subset also incurs lower inference cost, the even-cardinality ensemble cannot achieve higher utility under the surrogate objective. Consequently, considering only odd cardinalities does not exclude an optimal solution to the surrogate-based subset optimization problem. The greedy prefix construction further restricts the search space and does not guarantee the globally optimal subset, but provides an efficient procedure for selecting among candidate ensembles.  A comparison with exhaustive surrogate optimization is provided in Appendix~\ref{app:exact-vs-greedy}.

Once an ensemble $E$ has been selected, \routerM queries its constituent models in parallel and combines their responses through self-weighted BERTScore~\cite{Zhang*2020BERTScore}, a capability-aware extension of minimum Bayes risk (MBR) decoding. The resolver incorporates the estimated capabilities of both candidate and reference responses, weighting their contributions according to their predicted reliability. Let $y_i$ denote the response produced by model $i\in E$. We first normalize the capability estimates over the selected ensemble:
\begin{equation}
    w_i(x,E)
    =
    \frac{\hat p_i(x)}
         {\sum_{j\in E}\hat p_j(x)}.
\end{equation}
The score assigned to each candidate response is then
\begin{equation}
    S_i(x,E)
    =
    w_i(x,E)
    \sum_{j\in E}
    w_j(x,E)\operatorname{BERTScore}(y_i,y_j).
\end{equation}
The final ensemble response is selected as
\begin{equation}
    y^*
    =
    \arg\max_{i\in E}S_i(x,E).
\end{equation}
Alternative response resolution strategies are described in Appendix~\ref{app:experimental-setup}. Algorithm~\ref{alg:routerM} summarizes the complete inference procedure, from prompt-dependent capability estimation to cost-aware ensemble selection.

\begin{algorithm}[htb]
\caption{\routerM: Prompt-Adaptive Cost-Aware Ensembling}
\label{alg:routerM}
\begin{algorithmic}[1]
\Require query embedding $x$; trade-off parameter $\alpha$; maximum ensemble size $k_{\max}$
\Require fitted \router; average model costs $\{c_m^{\mathrm{avg}}\}_{m=1}^{M}$

\For{$m\in\mathcal{M}$}
    \State compute $\hat p_m(x)$ using \router
    \State compute $\bar C(\{m\})$ using Eq.~\eqref{eq:predicted-normalized-ensemble-cost}
    \State compute $J_{\{m\}}(x;\alpha)$ using Eq.~\eqref{eq:ensemble-ranking-utility}
\EndFor

\State sort models as $m_{(1)},\ldots,m_{(M)}$ such that
\[
    J_{\{m_{(1)}\}}(x;\alpha)
    \geq \cdots \geq
    J_{\{m_{(M)}\}}(x;\alpha)
\]

\State $\mathrm{best}\leftarrow\emptyset$
\State $\mathrm{bestval}\leftarrow-\infty$

\For{$k=1,\ldots,k_{\max}$ with $k \bmod 2 = 1$ }
    \State $E_k\leftarrow\{m_{(1)},\ldots,m_{(k)}\}$
    \State compute $\hat p_{E_k}^{\mathrm{maj}}(x)$
    \State compute $\bar C(E_k)$
    \State compute $J_{E_k}(x;\alpha)$ using Eq.~\eqref{eq:ensemble-utility}

    \If{$J_{E_k}(x;\alpha)>\mathrm{bestval}$}
        \State $\mathrm{best}\leftarrow E_k$
        \State $\mathrm{bestval}\leftarrow J_{E_k}(x;\alpha)$
    \EndIf
\EndFor

\State \Return $\mathrm{best}$
\end{algorithmic}
\end{algorithm}

\section{Experiments}
\label{sec:extension-gate}

\subsection{Experimental Setup}
\label{sec:experimental-setup}

\paragraph{Benchmarks.}
We evaluate \router{} and \routerM{} on four LLM routing benchmarks: LLMRouterBench Track-2~\cite{li2026LLMRouterBencha}, RouterBench~\cite{hu2024RouterBencha}, FusionBench~\cite{feng2026FusionFactory}, and SPROUT~\cite{somerstep2025CARROT}. These benchmarks cover heterogeneous candidate model pools and task groups, with performance measured using their respective evaluation scores.

\paragraph{Evaluation Protocol.}
We consider both ID and OOD evaluations. For ID experiments, we use the released data partitions when available; otherwise, we split the data into 70\% development and 30\% test, further dividing the development set into 80\% training and 20\% validation. For OOD experiments, we train exclusively on ID task groups, select configurations using ID validation data, and evaluate on held-out task groups. ID and OOD experiments use separately fitted routers. All experiments are evaluated over five seeds, with fixed data partitions, and all approaches are evaluated using the model responses provided by the benchmarks.

\paragraph{Baselines.} 

We compare \router{} against four single-model routing baselines: EmbedLLM~\cite{zhuang2025EmbedLLM}, Avengers-Pro~\cite{zhang2026Avengers,zhang2025GPT5}, K-NN, and a Multi-Layer Perceptron (MLP). For \routerM{}, we consider five LLM ensembling approaches: DFPE~\cite{cohen2026DFPE}, LLM-TOPLA~\cite{tekin2024LLMTOPLA}, SelectLLM~\cite{maurya2025SelectLLM}, PairRanker~\cite{jiang2023LLMBlender}, and BERTScore-based MBR~\cite{Zhang*2020BERTScore}. PairRanker employs PairRM, the pairwise reward model introduced by LLM-Blender, to rank candidate responses. We evaluate its pretrained checkpoint (PairRanker-PT) and a fine-tuned variant (PairRanker-FT), trained following the original procedure.

We additionally report an oracle single-model selector that uses ground-truth evaluation scores to select the best-performing model for each prompt, providing a reference for the performance attainable through perfect model selection.

\paragraph{Metrics.} We report task performance and inference cost, computed from benchmark-specific scores and model invocation costs. Task performance is macro-averaged across task groups. To summarize cost--performance trade-offs, we use normalized dominated hypervolume (HV), multiplied by 100. The reference performance corresponds to a random router, while the reference cost is set to the 40th percentile of the full-pool cost distribution for single-model routing and the 70th percentile for ensemble routing. These reference points are chosen to accommodate all the operating points of the compared methods. Since existing ensemble approaches generally provide a single operating point rather than a cost--performance frontier, we compare them directly in terms of task performance and inference cost.

\paragraph{Implementation.}
We employ Qwen3-Embedding-8B~\cite{zhang2025qwen3embeddingadvancingtext}  for creating the prompt embeddings required by the routers. \routerM{} considers ensembles of up to eleven models. Hyperparameters and ensemble configurations are selected independently for each benchmark using validation task performance at $\alpha=1$. The same fitted capability estimators are reused across different cost--performance operating points. Further implementation details, including the HV definition, are provided in Appendix~\ref{app:experimental-setup}.

\subsection{\router Evaluation}
\label{sec:router-evaluation}

\begin{table}[t]
\centering\scriptsize
\caption{Normalized hypervolume (HV) of single-model routers across ID and OOD benchmarks. HV is multiplied by 100 and uses the p40 full-pool cost reference. Parentheses report sample standard deviations across five seeds.}
\label{tab:router-hv}
\begin{tabularx}{\linewidth}{@{}l*{5}{>{\centering\arraybackslash}X}@{}}
\toprule
Approach & Track-2 & RouterBench & FusionBench & SPROUT & Macro average \\
\multicolumn{6}{c}{Normalized HV ($\times 100$; not percent)} \\
\midrule
\rowcolor{black!8}\multicolumn{6}{l}{\textbf{ID}} \\
EmbedLLM-MF & \ensuremath{\pmb{\mathbf{22.87}}}{\tiny\textcolor{black!45}{ (0.62)}} & \ensuremath{41.28}{\tiny\textcolor{black!45}{ (0.98)}} & \ensuremath{\underline{41.92}}{\tiny\textcolor{black!45}{ (1.32)}} & \ensuremath{\underline{37.03}}{\tiny\textcolor{black!45}{ (0.36)}} & \ensuremath{\underline{35.78}}{\tiny\textcolor{black!45}{ (0.51)}} \\
Avengers-Pro & \ensuremath{19.61}{\tiny\textcolor{black!45}{ (1.23)}} & \ensuremath{\underline{45.46}}{\tiny\textcolor{black!45}{ (0.81)}} & \ensuremath{40.46}{\tiny\textcolor{black!45}{ (1.57)}} & \ensuremath{36.22}{\tiny\textcolor{black!45}{ (0.96)}} & \ensuremath{35.44}{\tiny\textcolor{black!45}{ (0.60)}} \\
KNN & \ensuremath{19.80}{\tiny\textcolor{black!45}{ (0.00)}} & \ensuremath{43.03}{\tiny\textcolor{black!45}{ (0.00)}} & \ensuremath{39.93}{\tiny\textcolor{black!45}{ (0.00)}} & \ensuremath{34.74}{\tiny\textcolor{black!45}{ (0.00)}} & \ensuremath{34.38}{\tiny\textcolor{black!45}{ (0.00)}} \\
MLP & \ensuremath{1.79}{\tiny\textcolor{black!45}{ (1.24)}} & \ensuremath{27.99}{\tiny\textcolor{black!45}{ (0.31)}} & \ensuremath{33.95}{\tiny\textcolor{black!45}{ (0.89)}} & \ensuremath{22.34}{\tiny\textcolor{black!45}{ (1.47)}} & \ensuremath{21.52}{\tiny\textcolor{black!45}{ (0.69)}} \\
\router & \ensuremath{\underline{22.58}}{\tiny\textcolor{black!45}{ (0.42)}} & \ensuremath{\pmb{\mathbf{46.15}}}{\tiny\textcolor{black!45}{ (0.25)}} & \ensuremath{\pmb{\mathbf{42.57}}}{\tiny\textcolor{black!45}{ (0.98)}} & \ensuremath{\pmb{\mathbf{38.33}}}{\tiny\textcolor{black!45}{ (0.49)}} & \ensuremath{\pmb{\mathbf{37.41}}}{\tiny\textcolor{black!45}{ (0.35)}} \\
\midrule
\rowcolor{black!8}\multicolumn{6}{l}{\textbf{OOD}} \\
EmbedLLM-MF & \ensuremath{\underline{13.60}}{\tiny\textcolor{black!45}{ (1.32)}} & \ensuremath{27.90}{\tiny\textcolor{black!45}{ (0.36)}} & \ensuremath{\underline{19.74}}{\tiny\textcolor{black!45}{ (3.46)}} & \ensuremath{\underline{24.08}}{\tiny\textcolor{black!45}{ (0.25)}} & \ensuremath{\underline{21.33}}{\tiny\textcolor{black!45}{ (0.75)}} \\
Avengers-Pro & \ensuremath{3.48}{\tiny\textcolor{black!45}{ (3.85)}} & \ensuremath{\underline{28.46}}{\tiny\textcolor{black!45}{ (2.67)}} & \ensuremath{17.36}{\tiny\textcolor{black!45}{ (2.06)}} & \ensuremath{18.96}{\tiny\textcolor{black!45}{ (1.31)}} & \ensuremath{17.06}{\tiny\textcolor{black!45}{ (1.52)}} \\
KNN & \ensuremath{9.21}{\tiny\textcolor{black!45}{ (0.00)}} & \ensuremath{26.94}{\tiny\textcolor{black!45}{ (0.00)}} & \ensuremath{19.42}{\tiny\textcolor{black!45}{ (0.00)}} & \ensuremath{\pmb{\mathbf{24.53}}}{\tiny\textcolor{black!45}{ (0.00)}} & \ensuremath{20.02}{\tiny\textcolor{black!45}{ (0.00)}} \\
MLP & \ensuremath{2.66}{\tiny\textcolor{black!45}{ (0.88)}} & \ensuremath{15.10}{\tiny\textcolor{black!45}{ (4.55)}} & \ensuremath{18.19}{\tiny\textcolor{black!45}{ (2.79)}} & \ensuremath{14.38}{\tiny\textcolor{black!45}{ (0.90)}} & \ensuremath{12.58}{\tiny\textcolor{black!45}{ (1.01)}} \\
\router & \ensuremath{\pmb{\mathbf{17.18}}}{\tiny\textcolor{black!45}{ (1.85)}} & \ensuremath{\pmb{\mathbf{29.50}}}{\tiny\textcolor{black!45}{ (0.59)}} & \ensuremath{\pmb{\mathbf{22.91}}}{\tiny\textcolor{black!45}{ (1.21)}} & \ensuremath{21.81}{\tiny\textcolor{black!45}{ (0.65)}} & \ensuremath{\pmb{\mathbf{22.85}}}{\tiny\textcolor{black!45}{ (0.38)}} \\
\bottomrule
\end{tabularx}
\end{table}

We first evaluate \router as a single-model router, comparing its cost--performance trade-offs against four established routing baselines. 
Table~\ref{tab:router-hv} reports the HV across four benchmarks under ID and OOD conditions.

\router achieves the highest macro-average HV among the evaluated routers in both ID and OOD settings, reaching 37.41 and 22.85, respectively, compared with 35.78 and 21.33 for EmbedLLM-MF, the strongest competing router by this measure. PISA ranks first in six of the eight benchmark–setting combinations, showing that its aggregate advantage extends across multiple benchmarks. These results establish \router as an effective single-model router in its own right, while its capability estimates also provide the foundation for ensemble selection in \routerM.



\subsection{\routerM Evaluation}
\label{sec:main_results}

\begin{table*}[t]
\centering\scriptsize
\caption{ID test task performance (\%) and inference cost (USD/query $\times 10^{4}$) across four benchmarks and their macro average. Parentheses report sample standard deviations across five seeds where available.}
\label{tab:ensemble-id}
\begin{tabularx}{\textwidth}{l>{\centering\arraybackslash}X>{\centering\arraybackslash}X|>{\centering\arraybackslash}X>{\centering\arraybackslash}X|>{\centering\arraybackslash}X>{\centering\arraybackslash}X|>{\centering\arraybackslash}X>{\centering\arraybackslash}X|>{\centering\arraybackslash}X>{\centering\arraybackslash}X}
\toprule
Approach & \multicolumn{2}{c|}{Track-2} & \multicolumn{2}{c|}{RouterBench} & \multicolumn{2}{c|}{FusionBench} & \multicolumn{2}{c|}{SPROUT} & \multicolumn{2}{c}{Macro average} \\
\cmidrule(lr){2-3} \cmidrule(lr){4-5} \cmidrule(lr){6-7} \cmidrule(lr){8-9} \cmidrule(lr){10-11}
 & Perf. (\%) & Cost & Perf. (\%) & Cost & Perf. (\%) & Cost & Perf. (\%) & Cost & Perf. (\%) & Cost \\
\midrule
Oracle & \ensuremath{85.93} & \mbox{\ensuremath{213}} & \ensuremath{92.15} & \mbox{\ensuremath{13.7}} & \ensuremath{91.01} & \mbox{\ensuremath{4.48}} & \ensuremath{98.04} & \mbox{\ensuremath{62.2}} & \ensuremath{91.78} & \mbox{\ensuremath{73.3}} \\
\midrule
\rowcolor{black!8}\multicolumn{11}{l}{\textbf{Prior ensemble approaches}} \\
 PairRanker-PT & \ensuremath{60.87} & \mbox{\ensuremath{946}} & \ensuremath{67.71} & \mbox{\ensuremath{87}} & \ensuremath{51.03} & \mbox{\ensuremath{8.05}} & \ensuremath{68.54} & \mbox{\ensuremath{132}} & \ensuremath{62.04} & \mbox{\ensuremath{293}} \\
 PairRanker-FT & \ensuremath{62.18} & \mbox{\ensuremath{1200}} & \ensuremath{77.62} & \mbox{\ensuremath{87}} & \ensuremath{52.04} & \mbox{\ensuremath{8.05}} & \ensuremath{75.75} & \mbox{\ensuremath{162}} & \ensuremath{66.90} & \mbox{\ensuremath{365}} \\
MBR & \ensuremath{60.75} & \mbox{\ensuremath{1040}} & \ensuremath{70.22} & \mbox{\ensuremath{87}} & \ensuremath{62.40} & \mbox{\ensuremath{45.6}} & \ensuremath{71.31} & \mbox{\ensuremath{139}} & \ensuremath{66.17} & \mbox{\ensuremath{329}} \\
DFPE & \ensuremath{59.95} & \mbox{\ensuremath{965}} & \ensuremath{\underline{81.25}} & \mbox{\ensuremath{81.2}} & \ensuremath{69.17} & \mbox{\ensuremath{59.9}} & \ensuremath{76.30} & \mbox{\ensuremath{167}} & \ensuremath{71.67} & \mbox{\ensuremath{318}} \\
LLM-TOPLA & \ensuremath{57.93}{\tiny\textcolor{black!45}{ (0.43)}} & \mbox{\ensuremath{1050}{\tiny\textcolor{black!45}{ (99)}}} & \ensuremath{81.25}{\tiny\textcolor{black!45}{ (0.1)}} & \mbox{\ensuremath{119}{\tiny\textcolor{black!45}{ (33)}}} & \ensuremath{59.44}{\tiny\textcolor{black!45}{ (1.6)}} & \mbox{\ensuremath{15.5}{\tiny\textcolor{black!45}{ (8.5)}}} & \ensuremath{73.30}{\tiny\textcolor{black!45}{ (3.18)}} & \mbox{\ensuremath{18.4}{\tiny\textcolor{black!45}{ (8.9)}}} & \ensuremath{67.98}{\tiny\textcolor{black!45}{ (0.79)}} & \mbox{\ensuremath{301}{\tiny\textcolor{black!45}{ (26)}}} \\
SelectLLM & \ensuremath{61.55} & \mbox{\ensuremath{309}} & \ensuremath{78.93} & \mbox{\ensuremath{63.5}} & \ensuremath{74.62} & \mbox{\ensuremath{14.7}} & \ensuremath{\pmb{\mathbf{77.63}}} & \mbox{\ensuremath{142}} & \ensuremath{73.18} & \mbox{\ensuremath{132}} \\
\midrule
\rowcolor{black!8}\multicolumn{11}{l}{\textbf{Our approach: \routerM}} \\
\routerM{} ($\alpha=0.3$) & \ensuremath{59.86}{\tiny\textcolor{black!45}{ (0.24)}} & \mbox{\ensuremath{40}{\tiny\textcolor{black!45}{ (0.8)}}} & \ensuremath{74.53}{\tiny\textcolor{black!45}{ (0.35)}} & \mbox{\ensuremath{5.97}{\tiny\textcolor{black!45}{ (0.077)}}} & \ensuremath{71.61}{\tiny\textcolor{black!45}{ (0.56)}} & \mbox{\ensuremath{2.18}{\tiny\textcolor{black!45}{ (0.12)}}} & \ensuremath{77.01}{\tiny\textcolor{black!45}{ (0.24)}} & \mbox{\ensuremath{3.84}{\tiny\textcolor{black!45}{ (0.095)}}} & \ensuremath{70.75}{\tiny\textcolor{black!45}{ (0.18)}} & \mbox{\ensuremath{13}{\tiny\textcolor{black!45}{ (0.23)}}} \\
\routerM{} ($\alpha=0.6$) & \ensuremath{64.64}{\tiny\textcolor{black!45}{ (0.88)}} & \mbox{\ensuremath{175}{\tiny\textcolor{black!45}{ (11)}}} & \ensuremath{75.01}{\tiny\textcolor{black!45}{ (0.42)}} & \mbox{\ensuremath{12.9}{\tiny\textcolor{black!45}{ (1.5)}}} & \ensuremath{72.71}{\tiny\textcolor{black!45}{ (0.27)}} & \mbox{\ensuremath{4.14}{\tiny\textcolor{black!45}{ (0.31)}}} & \ensuremath{75.92}{\tiny\textcolor{black!45}{ (0.49)}} & \mbox{\ensuremath{11.5}{\tiny\textcolor{black!45}{ (0.41)}}} & \ensuremath{72.07}{\tiny\textcolor{black!45}{ (0.42)}} & \mbox{\ensuremath{50.8}{\tiny\textcolor{black!45}{ (3.1)}}} \\
\routerM{} ($\alpha=0.75$) & \ensuremath{66.18}{\tiny\textcolor{black!45}{ (0.33)}} & \mbox{\ensuremath{268}{\tiny\textcolor{black!45}{ (5.1)}}} & \ensuremath{78.64}{\tiny\textcolor{black!45}{ (0.35)}} & \mbox{\ensuremath{43.4}{\tiny\textcolor{black!45}{ (3.8)}}} & \ensuremath{73.47}{\tiny\textcolor{black!45}{ (0.27)}} & \mbox{\ensuremath{7.17}{\tiny\textcolor{black!45}{ (0.6)}}} & \ensuremath{76.86}{\tiny\textcolor{black!45}{ (0.29)}} & \mbox{\ensuremath{17.3}{\tiny\textcolor{black!45}{ (0.48)}}} & \ensuremath{73.78}{\tiny\textcolor{black!45}{ (0.16)}} & \mbox{\ensuremath{84}{\tiny\textcolor{black!45}{ (1.6)}}} \\
\routerM{} ($\alpha=0.85$) & \ensuremath{\underline{66.91}}{\tiny\textcolor{black!45}{ (0.57)}} & \mbox{\ensuremath{372}{\tiny\textcolor{black!45}{ (15)}}} & \ensuremath{80.43}{\tiny\textcolor{black!45}{ (0.29)}} & \mbox{\ensuremath{69.4}{\tiny\textcolor{black!45}{ (2.4)}}} & \ensuremath{\underline{74.95}}{\tiny\textcolor{black!45}{ (0.4)}} & \mbox{\ensuremath{14.5}{\tiny\textcolor{black!45}{ (1.6)}}} & \ensuremath{\underline{77.34}}{\tiny\textcolor{black!45}{ (0.33)}} & \mbox{\ensuremath{32}{\tiny\textcolor{black!45}{ (0.6)}}} & \underline{74.91}{\tiny\textcolor{black!45}{ (0.19)}} & \mbox{\ensuremath{122}{\tiny\textcolor{black!45}{ (4.1)}}} \\
\routerM{} ($\alpha=1$) & \ensuremath{\pmb{\mathbf{68.08}}}{\tiny\textcolor{black!45}{ (0.23)}} & \mbox{\ensuremath{774}{\tiny\textcolor{black!45}{ (31)}}} & \ensuremath{\pmb{\mathbf{81.26}}}{\tiny\textcolor{black!45}{ (0.1)}} & \mbox{\ensuremath{152}{\tiny\textcolor{black!45}{ (3.5)}}} & \ensuremath{\pmb{\mathbf{75.31}}}{\tiny\textcolor{black!45}{ (0.45)}} & \mbox{\ensuremath{34.1}{\tiny\textcolor{black!45}{ (1.2)}}} & \ensuremath{76.44}{\tiny\textcolor{black!45}{ (0.83)}} & \mbox{\ensuremath{159}{\tiny\textcolor{black!45}{ (1.6)}}} & \ensuremath{\pmb{\mathbf{75.27}}}{\tiny\textcolor{black!45}{ (0.27)}} & \mbox{\ensuremath{280}{\tiny\textcolor{black!45}{ (8.4)}}} \\
\bottomrule
\end{tabularx}
\end{table*}

\begin{table*}[t]
\centering\scriptsize
\caption{OOD test task performance (\%) and inference cost (USD/query $\times 10^{4}$) on held-out task groups across four benchmarks and their macro average. Parentheses report sample standard deviations across five seeds where available.}
\label{tab:ensemble-ood}
\begin{tabularx}{\textwidth}{l>{\centering\arraybackslash}X>{\centering\arraybackslash}X|>{\centering\arraybackslash}X>{\centering\arraybackslash}X|>{\centering\arraybackslash}X>{\centering\arraybackslash}X|>{\centering\arraybackslash}X>{\centering\arraybackslash}X|>{\centering\arraybackslash}X>{\centering\arraybackslash}X}
\toprule
Approach & \multicolumn{2}{c|}{Track-2} & \multicolumn{2}{c|}{RouterBench} & \multicolumn{2}{c|}{FusionBench} & \multicolumn{2}{c|}{SPROUT} & \multicolumn{2}{c}{Macro average} \\
\cmidrule(lr){2-3} \cmidrule(lr){4-5} \cmidrule(lr){6-7} \cmidrule(lr){8-9} \cmidrule(lr){10-11}
 & Perf. (\%) & Cost & Perf. (\%) & Cost & Perf. (\%) & Cost & Perf. (\%) & Cost & Perf. (\%) & Cost \\
\midrule
Oracle & \ensuremath{77.09} & \mbox{\ensuremath{303}} & \ensuremath{87.17} & \mbox{\ensuremath{12.3}} & \ensuremath{88.61} & \mbox{\ensuremath{3.97}} & \ensuremath{96.27} & \mbox{\ensuremath{50.1}} & \ensuremath{87.29} & \mbox{\ensuremath{92.3}} \\
\midrule
\rowcolor{black!8}\multicolumn{11}{l}{\textbf{Prior ensemble approaches}} \\
 PairRanker-PT & \ensuremath{56.94} & \mbox{\ensuremath{1920}} & \ensuremath{47.94} & \mbox{\ensuremath{82.6}} & \ensuremath{27.42} & \mbox{\ensuremath{8.73}} & \ensuremath{40.83} & \mbox{\ensuremath{122}} & \ensuremath{43.28} & \mbox{\ensuremath{534}} \\
 PairRanker-FT & \ensuremath{50.97} & \mbox{\ensuremath{1940}} & \ensuremath{66.57} & \mbox{\ensuremath{82.6}} & \ensuremath{26.81} & \mbox{\ensuremath{11.7}} & \ensuremath{65.27} & \mbox{\ensuremath{122}} & \ensuremath{52.40} & \mbox{\ensuremath{539}} \\
MBR & \ensuremath{52.96} & \mbox{\ensuremath{1840}} & \ensuremath{50.58} & \mbox{\ensuremath{82.6}} & \ensuremath{29.28} & \mbox{\ensuremath{41.5}} & \ensuremath{47.56} & \mbox{\ensuremath{135}} & \ensuremath{45.10} & \mbox{\ensuremath{524}} \\
DFPE & \ensuremath{52.98} & \mbox{\ensuremath{1230}} & \ensuremath{73.30} & \mbox{\ensuremath{78.9}} & \ensuremath{48.94} & \mbox{\ensuremath{69.3}} & \ensuremath{\pmb{\mathbf{66.25}}} & \mbox{\ensuremath{169}} & \ensuremath{60.37} & \mbox{\ensuremath{387}} \\
LLM-TOPLA & \ensuremath{53.55}{\tiny\textcolor{black!45}{ (4.49)}} & \mbox{\ensuremath{1240}{\tiny\textcolor{black!45}{ (620)}}} & \ensuremath{\underline{73.34}}{\tiny\textcolor{black!45}{ (0.31)}} & \mbox{\ensuremath{110}{\tiny\textcolor{black!45}{ (27)}}} & \ensuremath{36.06}{\tiny\textcolor{black!45}{ (3.81)}} & \mbox{\ensuremath{30.3}{\tiny\textcolor{black!45}{ (1.5)}}} & \ensuremath{61.80}{\tiny\textcolor{black!45}{ (4.06)}} & \mbox{\ensuremath{42.4}{\tiny\textcolor{black!45}{ (29)}}} & \ensuremath{56.18}{\tiny\textcolor{black!45}{ (1.27)}} & \mbox{\ensuremath{357}{\tiny\textcolor{black!45}{ (150)}}} \\
SelectLLM & \ensuremath{58.42} & \mbox{\ensuremath{787}} & \ensuremath{\pmb{\mathbf{73.46}}} & \mbox{\ensuremath{78.9}} & \ensuremath{39.85} & \mbox{\ensuremath{2.91}} & \ensuremath{64.02} & \mbox{\ensuremath{56.7}} & \ensuremath{58.94} & \mbox{\ensuremath{231}} \\
\midrule
\rowcolor{black!8}\multicolumn{11}{l}{\textbf{Our approach: \routerM}} \\
\routerM{} ($\alpha=0.3$) & \ensuremath{48.46}{\tiny\textcolor{black!45}{ (0.1)}} & \mbox{\ensuremath{21.4}{\tiny\textcolor{black!45}{ (0.98)}}} & \ensuremath{56.28} & \mbox{\ensuremath{3.64}} & \ensuremath{38.89}{\tiny\textcolor{black!45}{ (0.47)}} & \mbox{\ensuremath{2.63}{\tiny\textcolor{black!45}{ (0.073)}}} & \ensuremath{58.52}{\tiny\textcolor{black!45}{ (0.26)}} & \mbox{\ensuremath{4.37}{\tiny\textcolor{black!45}{ (0.054)}}} & \ensuremath{50.54}{\tiny\textcolor{black!45}{ (0.09)}} & \mbox{\ensuremath{8.01}{\tiny\textcolor{black!45}{ (0.25)}}} \\
\routerM{} ($\alpha=0.6$) & \ensuremath{55.42}{\tiny\textcolor{black!45}{ (0.13)}} & \mbox{\ensuremath{91.4}{\tiny\textcolor{black!45}{ (1.5)}}} & \ensuremath{56.28} & \mbox{\ensuremath{3.64}} & \ensuremath{44.93}{\tiny\textcolor{black!45}{ (0.45)}} & \mbox{\ensuremath{6.14}{\tiny\textcolor{black!45}{ (0.2)}}} & \ensuremath{64.69}{\tiny\textcolor{black!45}{ (0.23)}} & \mbox{\ensuremath{19.4}{\tiny\textcolor{black!45}{ (0.26)}}} & \ensuremath{55.33}{\tiny\textcolor{black!45}{ (0.15)}} & \mbox{\ensuremath{30.1}{\tiny\textcolor{black!45}{ (0.42)}}} \\
\routerM{} ($\alpha=0.75$) & \ensuremath{58.57}{\tiny\textcolor{black!45}{ (0.58)}} & \mbox{\ensuremath{375}{\tiny\textcolor{black!45}{ (74)}}} & \ensuremath{56.42}{\tiny\textcolor{black!45}{ (0.15)}} & \mbox{\ensuremath{3.69}{\tiny\textcolor{black!45}{ (0.038)}}} & \ensuremath{49.37}{\tiny\textcolor{black!45}{ (0.79)}} & \mbox{\ensuremath{11.8}{\tiny\textcolor{black!45}{ (0.23)}}} & \ensuremath{63.03}{\tiny\textcolor{black!45}{ (0.27)}} & \mbox{\ensuremath{49}{\tiny\textcolor{black!45}{ (2.5)}}} & \ensuremath{56.85}{\tiny\textcolor{black!45}{ (0.31)}} & \mbox{\ensuremath{110}{\tiny\textcolor{black!45}{ (19)}}} \\
\routerM{} ($\alpha=0.85$) & \ensuremath{\underline{61.81}}{\tiny\textcolor{black!45}{ (0.61)}} & \mbox{\ensuremath{996}{\tiny\textcolor{black!45}{ (65)}}} & \ensuremath{66.94}{\tiny\textcolor{black!45}{ (2.06)}} & \mbox{\ensuremath{69.8}{\tiny\textcolor{black!45}{ (8.2)}}} & \ensuremath{\underline{50.10}}{\tiny\textcolor{black!45}{ (0.4)}} & \mbox{\ensuremath{16.4}{\tiny\textcolor{black!45}{ (0.34)}}} & \ensuremath{64.56}{\tiny\textcolor{black!45}{ (0.32)}} & \mbox{\ensuremath{87.2}{\tiny\textcolor{black!45}{ (1.7)}}} & \underline{60.85}{\tiny\textcolor{black!45}{ (0.65)}} & \mbox{\ensuremath{292}{\tiny\textcolor{black!45}{ (18)}}} \\
\routerM{} ($\alpha=1$) & \ensuremath{\pmb{\mathbf{62.65}}}{\tiny\textcolor{black!45}{ (0.25)}} & \mbox{\ensuremath{3100}{\tiny\textcolor{black!45}{ (20)}}} & \ensuremath{72.79}{\tiny\textcolor{black!45}{ (0.6)}} & \mbox{\ensuremath{115}{\tiny\textcolor{black!45}{ (15)}}} & \ensuremath{\pmb{\mathbf{52.16}}}{\tiny\textcolor{black!45}{ (0.21)}} & \mbox{\ensuremath{44.8}{\tiny\textcolor{black!45}{ (0.6)}}} & \ensuremath{\underline{65.34}}{\tiny\textcolor{black!45}{ (0.26)}} & \mbox{\ensuremath{159}{\tiny\textcolor{black!45}{ (1.3)}}} & \ensuremath{\pmb{\mathbf{63.24}}}{\tiny\textcolor{black!45}{ (0.15)}} & \mbox{\ensuremath{854}{\tiny\textcolor{black!45}{ (6.8)}}} \\
\bottomrule
\end{tabularx}
\end{table*}

\begin{table}[!t]
\centering\scriptsize
\caption{Ensemble size, performance, and cost of \routerM{} versus $\alpha$. Singleton (\%) denotes queries using one model. ``Both'' averages all eight scenarios; parentheses give sample standard deviations over five seeds.}
\label{tab:ensemble-cardinality}
\begin{tabular}{lrrrrr}
\toprule
Split & $\alpha$ & Models/query & Singleton (\%) & Perf. (\%) & Cost (USD/query $\times10^{4}$) \\
\midrule
ID & 0.3 & \ensuremath{1.32}{\tiny\textcolor{black!45}{ (0.05)}} & \ensuremath{84.33}{\tiny\textcolor{black!45}{ (2.46)}} & \ensuremath{70.75}{\tiny\textcolor{black!45}{ (0.18)}} & \mbox{\ensuremath{\pmb{\mathbf{13}}}{\tiny\textcolor{black!45}{ (0.23)}}} \\
ID & 0.6 & \ensuremath{2.00}{\tiny\textcolor{black!45}{ (0.04)}} & \ensuremath{55.70}{\tiny\textcolor{black!45}{ (2.02)}} & \ensuremath{72.07}{\tiny\textcolor{black!45}{ (0.42)}} & \mbox{\ensuremath{50.8}{\tiny\textcolor{black!45}{ (3.1)}}} \\
ID & 1.0 & \ensuremath{6.42}{\tiny\textcolor{black!45}{ (0.15)}} & \ensuremath{21.56}{\tiny\textcolor{black!45}{ (1.39)}} & \ensuremath{\pmb{\mathbf{75.27}}}{\tiny\textcolor{black!45}{ (0.27)}} & \mbox{\ensuremath{280}{\tiny\textcolor{black!45}{ (8.4)}}} \\
\midrule
OOD & 0.3 & \ensuremath{1.35}{\tiny\textcolor{black!45}{ (0.01)}} & \ensuremath{82.41}{\tiny\textcolor{black!45}{ (0.62)}} & \ensuremath{50.54}{\tiny\textcolor{black!45}{ (0.09)}} & \mbox{\ensuremath{\pmb{\mathbf{8.01}}}{\tiny\textcolor{black!45}{ (0.25)}}} \\
OOD & 0.6 & \ensuremath{2.41}{\tiny\textcolor{black!45}{ (0.02)}} & \ensuremath{37.44}{\tiny\textcolor{black!45}{ (1.23)}} & \ensuremath{55.33}{\tiny\textcolor{black!45}{ (0.15)}} & \mbox{\ensuremath{30.1}{\tiny\textcolor{black!45}{ (0.42)}}} \\
OOD & 1.0 & \ensuremath{7.39}{\tiny\textcolor{black!45}{ (0.15)}} & \ensuremath{9.31}{\tiny\textcolor{black!45}{ (0.09)}} & \ensuremath{\pmb{\mathbf{63.24}}}{\tiny\textcolor{black!45}{ (0.15)}} & \mbox{\ensuremath{854}{\tiny\textcolor{black!45}{ (6.8)}}} \\
\midrule
Both & 0.3 & \ensuremath{1.34}{\tiny\textcolor{black!45}{ (0.03)}} & \ensuremath{83.37}{\tiny\textcolor{black!45}{ (1.44)}} & \ensuremath{60.64}{\tiny\textcolor{black!45}{ (0.12)}} & \mbox{\ensuremath{\pmb{\mathbf{10.5}}}{\tiny\textcolor{black!45}{ (0.13)}}} \\
Both & 0.6 & \ensuremath{2.21}{\tiny\textcolor{black!45}{ (0.02)}} & \ensuremath{46.57}{\tiny\textcolor{black!45}{ (0.82)}} & \ensuremath{63.70}{\tiny\textcolor{black!45}{ (0.21)}} & \mbox{\ensuremath{40.4}{\tiny\textcolor{black!45}{ (1.6)}}} \\
Both & 1.0 & \ensuremath{6.91}{\tiny\textcolor{black!45}{ (0.14)}} & \ensuremath{15.43}{\tiny\textcolor{black!45}{ (0.68)}} & \ensuremath{\pmb{\mathbf{69.25}}}{\tiny\textcolor{black!45}{ (0.17)}} & \mbox{\ensuremath{567}{\tiny\textcolor{black!45}{ (5.7)}}} \\
\bottomrule
\end{tabular}
\end{table}

We evaluate \routerM against existing LLM ensembling approaches across the four benchmarks. Tables~\ref{tab:ensemble-id} and~\ref{tab:ensemble-ood} report task performance and inference cost under ID and OOD conditions, respectively.

\textbf{\routerM{} attains higher macro-average task performance than the strongest competing ensemble approaches while requiring less inference expenditure.} On ID benchmarks, \routerM{} at $\alpha=0.75$ achieves 73.78\% task performance at a cost of $84 \times 10^{-4}$ USD/query, compared with 73.18\% at $132 \times 10^{-4}$ USD/query of the best competitor, SelectLLM. This represents a \textbf{0.60 percentage-point improvement with approximately 36\% lower cost}. Under OOD conditions, \routerM{} at $\alpha=0.85$ achieves 60.85\% performance at a cost of $292 \times 10^{-4}$ USD/query, compared with 60.37\% at $387 \times 10^{-4}$ USD/query for DFPE, \textbf{improving performance by 0.48 percentage points while reducing cost by approximately 25\%}.

When response quality is prioritized ($\alpha=1$), \textbf{\routerM reaches the highest macro-average task performance among the evaluated deployable approaches in both settings}, reaching 75.27\% on ID benchmarks and 63.24\% on OOD benchmarks. These scores exceed those of the strongest competing ensemble approaches by \textbf{2.09 and 2.87 percentage points}, respectively. Although some competitors achieve higher performance on individual benchmarks, none matches \routerM in macro-average task performance across either evaluation setting.

Thus, \routerM supports different cost--performance preferences by varying the trade-off parameter $\alpha$. On ID benchmarks, decreasing $\alpha$ from 1.0 to 0.3 reduces average inference cost by \textbf{a factor of 21.5}, while task performance decreases from 75.27\% to 70.75\%. On OOD benchmarks, the corresponding cost reduction is approximately \textbf{106-fold}, with performance decreasing from 63.24\% to 50.54\%. 


Table~\ref{tab:ensemble-cardinality} reports how ensemble cardinality varies with the cost preference $\alpha$, together with the proportion of queries assigned to a single model. At $\alpha=0.3$, \routerM selects a single model for 84.33\% of ID queries and 82.41\% of OOD queries. Increasing $\alpha$ leads to larger ensembles on average, with cardinalities reaching 6.42 and 7.39 at $\alpha=1.0$ under ID and OOD conditions, respectively. Even when inference cost is not penalized, however, \routerM selects a single model for 21.56\% of ID queries and 9.31\% of OOD queries. \textbf{Ensemble cardinality therefore remains prompt-dependent even when the selection objective prioritizes response quality exclusively}.


\subsection{Adaptive vs. Fixed Ensembling}
\label{sec:adaptive-vs-fixed}

\begin{table*}[t]
\centering\scriptsize
\caption{Normalized HV ($\times 100$) of adaptive and fixed-cardinality ensemble routing under ID and OOD conditions, using the p70 cost reference. Parentheses report sample standard deviations across five seeds.}
\label{tab:adaptive-vs-fixed}
\begin{tabularx}{\textwidth}{l*{7}{>{\centering\arraybackslash}X}}
\toprule
Scenario & \routerM & Fixed $k=1$ & Fixed $k=3$ & Fixed $k=5$ & Fixed $k=7$ & Fixed $k=9$ & Fixed $k=11$ \\
\midrule
\rowcolor{black!8}\multicolumn{8}{l}{\textbf{ID}} \\
Track-2 & \ensuremath{\pmb{\mathbf{28.57}}}{\tiny\textcolor{black!45}{ (0.59)}} & \ensuremath{\underline{27.89}}{\tiny\textcolor{black!45}{ (0.49)}} & \ensuremath{27.80}{\tiny\textcolor{black!45}{ (0.53)}} & \ensuremath{26.69}{\tiny\textcolor{black!45}{ (0.26)}} & \ensuremath{24.06}{\tiny\textcolor{black!45}{ (0.38)}} & \ensuremath{19.90}{\tiny\textcolor{black!45}{ (0.41)}} & \ensuremath{14.55}{\tiny\textcolor{black!45}{ (0.40)}} \\
RouterBench & \ensuremath{\underline{46.37}}{\tiny\textcolor{black!45}{ (0.31)}} & \ensuremath{\pmb{\mathbf{46.99}}}{\tiny\textcolor{black!45}{ (0.25)}} & \ensuremath{44.71}{\tiny\textcolor{black!45}{ (0.30)}} & \ensuremath{41.66}{\tiny\textcolor{black!45}{ (0.38)}} & \ensuremath{38.75}{\tiny\textcolor{black!45}{ (0.37)}} & \ensuremath{28.95}{\tiny\textcolor{black!45}{ (0.21)}} & \ensuremath{2.67}{\tiny\textcolor{black!45}{ (0.02)}} \\
FusionBench & \ensuremath{\pmb{\mathbf{44.89}}}{\tiny\textcolor{black!45}{ (0.87)}} & \ensuremath{42.74}{\tiny\textcolor{black!45}{ (0.98)}} & \ensuremath{\underline{44.72}}{\tiny\textcolor{black!45}{ (0.75)}} & \ensuremath{44.42}{\tiny\textcolor{black!45}{ (0.66)}} & \ensuremath{43.74}{\tiny\textcolor{black!45}{ (0.64)}} & \ensuremath{42.90}{\tiny\textcolor{black!45}{ (0.89)}} & \ensuremath{41.60}{\tiny\textcolor{black!45}{ (0.60)}} \\
SPROUT & \ensuremath{\underline{37.57}}{\tiny\textcolor{black!45}{ (0.67)}} & \ensuremath{\pmb{\mathbf{38.65}}}{\tiny\textcolor{black!45}{ (0.49)}} & \ensuremath{36.13}{\tiny\textcolor{black!45}{ (1.13)}} & \ensuremath{35.86}{\tiny\textcolor{black!45}{ (0.89)}} & \ensuremath{34.27}{\tiny\textcolor{black!45}{ (0.92)}} & \ensuremath{31.83}{\tiny\textcolor{black!45}{ (0.45)}} & \ensuremath{27.61}{\tiny\textcolor{black!45}{ (0.60)}} \\
ID macro average & \ensuremath{\pmb{\mathbf{39.35}}}{\tiny\textcolor{black!45}{ (0.44)}} & \ensuremath{\underline{39.07}}{\tiny\textcolor{black!45}{ (0.37)}} & \ensuremath{38.34}{\tiny\textcolor{black!45}{ (0.52)}} & \ensuremath{37.16}{\tiny\textcolor{black!45}{ (0.40)}} & \ensuremath{35.21}{\tiny\textcolor{black!45}{ (0.42)}} & \ensuremath{30.89}{\tiny\textcolor{black!45}{ (0.36)}} & \ensuremath{21.61}{\tiny\textcolor{black!45}{ (0.29)}} \\
\midrule
\rowcolor{black!8}\multicolumn{8}{l}{\textbf{OOD}} \\
Track-2 & \ensuremath{\underline{22.39}}{\tiny\textcolor{black!45}{ (0.34)}} & \ensuremath{20.54}{\tiny\textcolor{black!45}{ (1.67)}} & \ensuremath{21.69}{\tiny\textcolor{black!45}{ (0.76)}} & \ensuremath{\pmb{\mathbf{23.25}}}{\tiny\textcolor{black!45}{ (0.32)}} & \ensuremath{20.35}{\tiny\textcolor{black!45}{ (0.08)}} & \ensuremath{17.51}{\tiny\textcolor{black!45}{ (0.20)}} & \ensuremath{13.56}{\tiny\textcolor{black!45}{ (0.39)}} \\
RouterBench & \ensuremath{\underline{26.86}}{\tiny\textcolor{black!45}{ (1.29)}} & \ensuremath{\pmb{\mathbf{31.09}}}{\tiny\textcolor{black!45}{ (0.62)}} & \ensuremath{24.13}{\tiny\textcolor{black!45}{ (0.42)}} & \ensuremath{19.28}{\tiny\textcolor{black!45}{ (1.40)}} & \ensuremath{18.27}{\tiny\textcolor{black!45}{ (0.87)}} & \ensuremath{16.32}{\tiny\textcolor{black!45}{ (1.18)}} & \ensuremath{5.72}{\tiny\textcolor{black!45}{ (0.15)}} \\
FusionBench & \ensuremath{\pmb{\mathbf{30.96}}}{\tiny\textcolor{black!45}{ (0.22)}} & \ensuremath{23.03}{\tiny\textcolor{black!45}{ (1.23)}} & \ensuremath{26.42}{\tiny\textcolor{black!45}{ (0.37)}} & \ensuremath{27.61}{\tiny\textcolor{black!45}{ (0.28)}} & \ensuremath{29.16}{\tiny\textcolor{black!45}{ (0.23)}} & \ensuremath{29.95}{\tiny\textcolor{black!45}{ (0.24)}} & \ensuremath{\underline{30.11}}{\tiny\textcolor{black!45}{ (0.20)}} \\
SPROUT & \ensuremath{\pmb{\mathbf{24.98}}}{\tiny\textcolor{black!45}{ (0.44)}} & \ensuremath{23.32}{\tiny\textcolor{black!45}{ (0.64)}} & \ensuremath{23.68}{\tiny\textcolor{black!45}{ (0.32)}} & \ensuremath{\underline{23.70}}{\tiny\textcolor{black!45}{ (0.53)}} & \ensuremath{22.59}{\tiny\textcolor{black!45}{ (0.82)}} & \ensuremath{22.09}{\tiny\textcolor{black!45}{ (0.50)}} & \ensuremath{17.64}{\tiny\textcolor{black!45}{ (0.36)}} \\
OOD macro average & \ensuremath{\pmb{\mathbf{26.30}}}{\tiny\textcolor{black!45}{ (0.27)}} & \ensuremath{\underline{24.50}}{\tiny\textcolor{black!45}{ (0.31)}} & \ensuremath{23.98}{\tiny\textcolor{black!45}{ (0.29)}} & \ensuremath{23.46}{\tiny\textcolor{black!45}{ (0.43)}} & \ensuremath{22.59}{\tiny\textcolor{black!45}{ (0.14)}} & \ensuremath{21.47}{\tiny\textcolor{black!45}{ (0.17)}} & \ensuremath{16.76}{\tiny\textcolor{black!45}{ (0.15)}} \\
\midrule
\rowcolor{black!8}\multicolumn{8}{l}{\textbf{All scenarios}} \\
Full macro average & \ensuremath{\pmb{\mathbf{32.82}}}{\tiny\textcolor{black!45}{ (0.25)}} & \ensuremath{\underline{31.78}}{\tiny\textcolor{black!45}{ (0.31)}} & \ensuremath{31.16}{\tiny\textcolor{black!45}{ (0.35)}} & \ensuremath{30.31}{\tiny\textcolor{black!45}{ (0.37)}} & \ensuremath{28.90}{\tiny\textcolor{black!45}{ (0.27)}} & \ensuremath{26.18}{\tiny\textcolor{black!45}{ (0.21)}} & \ensuremath{19.18}{\tiny\textcolor{black!45}{ (0.17)}} \\
\bottomrule
\end{tabularx}
\end{table*}

We investigate whether prompt-adaptive ensemble cardinality improves cost--performance trade-offs over fixed-cardinality alternatives. To isolate this effect, we compare \routerM against variants of the same framework constrained to select exactly $k$ models per query, with $k \in \{1,3,5,7,9,11\}$. In particular, $k=1$ corresponds to \router. All variants use the same capability estimates, ensemble-success surrogate, and resolver, while varying $\alpha$ to obtain different operating points.

Table~\ref{tab:adaptive-vs-fixed} reports the HV achieved by adaptive and fixed-cardinality selection under ID and OOD conditions. \textbf{\routerM{} achieves the highest macro-average HV in both settings}, reaching 39.35 on ID benchmarks and 26.30 on OOD benchmarks. It also outperforms the strongest fixed-cardinality alternatives, with improvements of 0.28 HV points under ID conditions and \textbf{1.80 points under OOD conditions}. These results are particularly noteworthy because the competing configurations are optimized separately for each benchmark, whereas \routerM{} determines ensemble cardinality dynamically for individual prompts at inference time. The larger advantage under OOD conditions further highlights the benefits of adaptive cardinality when generalizing to previously unseen task groups.

The results also reveal differences across benchmarks. While adaptive cardinality improves HV in several cases, fixed-cardinality strategies remain competitive and occasionally outperform \routerM. In particular, fixed $k=1$ achieves higher HV on RouterBench in both settings, indicating that invoking additional models does not necessarily improve the trade-off for every task distribution. Nevertheless, the aggregate results suggest that ensemble cardinality is better treated as a prompt-dependent routing decision rather than a fixed configuration parameter, allowing additional model invocations to be allocated selectively according to predicted utility and inference-cost preferences.

\subsection{Component Analysis}
\label{sec:component-analysis}

\begin{table}[t]
\centering\scriptsize
\caption{Task performance (\%) at $\alpha=1$ for different ensemble-success estimators and response resolvers. The ensemble-success estimators are Poisson-binomial (PB) and Gaussian copula (GC). ID and OOD results are macro-averaged across the four benchmarks, while ``Both'' averages all eight scenarios. Parentheses report sample standard deviations across five seeds.}
\label{tab:component-analysis}
\begin{tabular}{llrrr}
\toprule
Selector & Resolver & ID & OOD & Both \\
\midrule
PB & Self-weighted BERTScore & \ensuremath{\pmb{\mathbf{75.27}}}{\tiny\textcolor{black!45}{ (0.27)}} & \ensuremath{\pmb{\mathbf{63.24}}}{\tiny\textcolor{black!45}{ (0.15)}} & \ensuremath{\pmb{\mathbf{69.25}}}{\tiny\textcolor{black!45}{ (0.17)}} \\
PB & Other-weighted BERTScore & \ensuremath{73.04}{\tiny\textcolor{black!45}{ (0.26)}} & \ensuremath{59.88}{\tiny\textcolor{black!45}{ (0.26)}} & \ensuremath{66.46}{\tiny\textcolor{black!45}{ (0.21)}} \\
PB & BERTScore & \ensuremath{74.35}{\tiny\textcolor{black!45}{ (0.19)}} & \ensuremath{62.03}{\tiny\textcolor{black!45}{ (0.15)}} & \ensuremath{68.19}{\tiny\textcolor{black!45}{ (0.06)}} \\
PB & PairRanker (pretrained) & \ensuremath{68.00}{\tiny\textcolor{black!45}{ (0.48)}} & \ensuremath{48.68}{\tiny\textcolor{black!45}{ (0.17)}} & \ensuremath{58.34}{\tiny\textcolor{black!45}{ (0.23)}} \\
PB & PairRanker (fine-tuned) & \ensuremath{71.65}{\tiny\textcolor{black!45}{ (0.31)}} & \ensuremath{52.07}{\tiny\textcolor{black!45}{ (0.24)}} & \ensuremath{61.86}{\tiny\textcolor{black!45}{ (0.19)}} \\
\midrule
GC & Self-weighted BERTScore & \ensuremath{75.23}{\tiny\textcolor{black!45}{ (0.13)}} & \ensuremath{62.36}{\tiny\textcolor{black!45}{ (0.22)}} & \ensuremath{68.80}{\tiny\textcolor{black!45}{ (0.13)}} \\
GC & Other-weighted BERTScore & \ensuremath{74.99}{\tiny\textcolor{black!45}{ (0.30)}} & \ensuremath{60.99}{\tiny\textcolor{black!45}{ (0.35)}} & \ensuremath{67.99}{\tiny\textcolor{black!45}{ (0.28)}} \\
GC & BERTScore & \ensuremath{75.18}{\tiny\textcolor{black!45}{ (0.25)}} & \ensuremath{61.91}{\tiny\textcolor{black!45}{ (0.22)}} & \ensuremath{68.54}{\tiny\textcolor{black!45}{ (0.18)}} \\
GC & PairRanker (pretrained) & \ensuremath{74.43}{\tiny\textcolor{black!45}{ (0.34)}} & \ensuremath{57.07}{\tiny\textcolor{black!45}{ (1.11)}} & \ensuremath{65.75}{\tiny\textcolor{black!45}{ (0.61)}} \\
GC & PairRanker (fine-tuned) & \ensuremath{74.95}{\tiny\textcolor{black!45}{ (0.15)}} & \ensuremath{57.94}{\tiny\textcolor{black!45}{ (0.53)}} & \ensuremath{66.44}{\tiny\textcolor{black!45}{ (0.30)}} \\
\bottomrule
\end{tabular}
\end{table}

We analyze two components of \routerM: the ensemble-success estimator used to select model subsets and the response resolver used to combine their outputs. For ensemble-success estimation, we compare the default Poisson-binomial (PB) formulation with a Gaussian-copula (GC) alternative that explicitly models dependencies among model correctness events. For response resolution, we consider standard BERTScore-based minimum Bayes risk (MBR) decoding, capability-weighted BERTScore variants, and pretrained or fine-tuned PairRanker. Standard BERTScore selects the response with the highest aggregate semantic similarity to the other ensemble responses. Other-weighted BERTScore weights the reference responses according to their estimated model capabilities, while self-weighted BERTScore additionally weights each candidate response by its own estimated capability. PairRanker instead employs a learned pairwise response-ranking model. Further details on the alternative estimator and resolvers are provided in the appendix.

\paragraph{Ensemble-Success Estimation.}
We investigate the impact of the ensemble-success estimator by comparing PB and GC across different response resolvers. Table~\ref{tab:component-analysis} reports task performance at $\alpha=1$ under ID and OOD conditions. GC improves performance for several resolver configurations, particularly other-weighted BERTScore and PairRanker. However, these improvements do not consistently extend to self-weighted BERTScore, for which PB achieves higher macro-average performance in both ID (75.27\% versus 75.23\%) and OOD (63.24\% versus 62.36\%) settings. Overall, \textbf{PB combined with self-weighted BERTScore achieves the highest aggregate performance}, indicating that explicitly modeling correctness dependencies does not consistently improve ensemble performance and supporting the simpler PB formulation as the default estimator.

\paragraph{Ensemble Resolver Quality.}

We further investigate how the choice of response resolver affects ensemble performance. As shown in Table~\ref{tab:component-analysis}, self-weighted BERTScore achieves the highest aggregate performance among the evaluated resolvers when paired with PB, outperforming standard BERTScore by 0.92 percentage points in ID and 1.21 percentage points in OOD settings. In contrast, PairRanker exhibits greater sensitivity to the ensemble-success estimator, with GC substantially improving performance over PB for both pretrained and fine-tuned variants. Fine-tuning consistently improves PairRanker performance over its pretrained counterpart under both estimation methods, although neither variant surpasses self-weighted BERTScore in aggregate performance. These findings support capability-weighted semantic similarity as an effective response resolution strategy and motivate self-weighted BERTScore as the default resolver in \routerM.

\subsection{Ablation of Prompt-Dependent Model Ranking}
\label{app:fixed-global-adaptive}

\begin{table}[t]
\centering\scriptsize
\caption{Effect of prompt-dependent model ranking on normalized hypervolume (HV, $\times 100$) under ID and OOD conditions, using the p70 cost reference. \routerM{} is compared with a variant that ranks models globally by their task performance on the fit set while retaining Poisson-binomial (PB) ensemble-success estimation.}
\label{tab:global-ranking-ablation}
\begin{tabularx}{\linewidth}{l*{2}{>{\centering\arraybackslash}X}}
\toprule
Scenario & \routerM & Fixed global + PB \\
\midrule
\rowcolor{black!8}\multicolumn{3}{l}{\textbf{ID}} \\
Track-2 & \ensuremath{\pmb{\mathbf{28.57}}}{\tiny\textcolor{black!45}{ (0.59)}} & \ensuremath{ {22.86}} \\
RouterBench & \ensuremath{\pmb{\mathbf{46.37}}}{\tiny\textcolor{black!45}{ (0.31)}} & \ensuremath{ {39.12}} \\
FusionBench & \ensuremath{\pmb{\mathbf{44.89}}}{\tiny\textcolor{black!45}{ (0.87)}} & \ensuremath{ {41.26}} \\
SPROUT & \ensuremath{\pmb{\mathbf{37.57}}}{\tiny\textcolor{black!45}{ (0.67)}} & \ensuremath{ {34.31}} \\
ID macro average & \ensuremath{\pmb{\mathbf{39.35}}}{\tiny\textcolor{black!45}{ (0.44)}} & \ensuremath{ {34.39}} \\
\midrule
\rowcolor{black!8}\multicolumn{3}{l}{\textbf{OOD}} \\
Track-2 & \ensuremath{\pmb{\mathbf{22.39}}}{\tiny\textcolor{black!45}{ (0.34)}} & \ensuremath{ {14.95}} \\
RouterBench & \ensuremath{\pmb{\mathbf{26.86}}}{\tiny\textcolor{black!45}{ (1.29)}} & \ensuremath{ {24.90}} \\
FusionBench & \ensuremath{ {30.96}}{\tiny\textcolor{black!45}{ (0.22)}} & \ensuremath{\pmb{\mathbf{34.91}}} \\
SPROUT & \ensuremath{ {24.98}}{\tiny\textcolor{black!45}{ (0.44)}} & \ensuremath{\pmb{\mathbf{26.81}}} \\
OOD macro average & \ensuremath{\pmb{\mathbf{26.30}}}{\tiny\textcolor{black!45}{ (0.27)}} & \ensuremath{ {25.39}} \\
\midrule
\rowcolor{black!8}\multicolumn{3}{l}{\textbf{All scenarios}} \\
Full macro average & \ensuremath{\pmb{\mathbf{32.82}}}{\tiny\textcolor{black!45}{ (0.25)}} & \ensuremath{ {29.89}} \\
\bottomrule
\end{tabularx}
\end{table}

To assess the contribution of prompt-dependent model ranking, we compare \routerM{} against a fixed-global baseline that retains adaptive ensemble cardinality but ranks models using their global training-set capabilities instead of prompt-specific estimates. This isolates the effect of prompt-dependent ranking while preserving the cost-aware ensemble selection framework. Table~\ref{tab:global-ranking-ablation} shows that \routerM{} improves macro-average HV by \textbf{4.96 points under ID conditions and 0.91 points under OOD conditions}. It outperforms the fixed-global baseline on all four ID benchmarks and on Track-2 and RouterBench under OOD conditions, while the baseline performs better on FusionBench and SPROUT. Across all eight scenarios, \routerM{} achieves a macro-average HV of 32.82 compared with 29.89 for the fixed-global baseline, indicating that prompt-dependent ranking contributes to cost--performance improvements independently of adaptive cardinality.

\section{Related Work}

\paragraph{Cost-Aware Model Routing}
LLM routing selects a model for each prompt by estimating response quality and accounting for inference cost. EmbedLLM~\cite{zhuang2025EmbedLLM} learns latent representations from a model--question correctness matrix. Clustering-based methods instead use similarities among prompts to estimate model performance: UniRoute~\cite{jitkrittum2026Universal} and Avengers-PRO~\cite{zhang2025GPT5,zhang2026Avengers} rely on hard cluster assignments and cluster-level statistics, while ZeroRouter~\cite{yan2026Breakingb} combines clustering with item response theory. PISA differs in using soft cluster memberships and model-specific predictors to estimate prompt-dependent capabilities. Related cost-aware inference strategies include LLM cascades~\cite{chen2024FrugalGPT,dekoninck2025Unified} and risk-aware routing~\cite{zhang2026RACER}. Cascades may escalate requests sequentially; PISA-E instead selects models for parallel execution before generating their responses.

\paragraph{LLM Ensembling and Response Aggregation}
LLM ensembling combines responses from multiple models to exploit complementary capabilities. Aggregation methods include majority voting~\cite{wang2023selfconsistency,ai2026Majority}, learned ranking and fusion, as in LLM-Blender~\cite{jiang2023LLMBlender} and URG~\cite{lv2024URG}, and similarity-based minimum Bayes risk decoding, including variants that account for uncertainty~\cite{daheim2025UncertaintyAware}. Multi-agent debate~\cite{du2024Improvinga} and Mixture-of-Agents~\cite{wang2025MOA} also combine model outputs, but involve interaction across successive rounds or stages. PISA-E does not organize such collaboration: after selecting an ensemble, it queries the models in parallel and resolves their responses using capability-weighted semantic similarity.

\paragraph{Adaptive and Cost-Aware Ensemble Selection}
Ensemble selection determines which models to invoke, and whether that choice changes across queries. LLM-TOPLA~\cite{tekin2024LLMTOPLA} and DFPE~\cite{cohen2026DFPE} construct ensembles that are subsequently reused across queries, whereas SelectLLM~\cite{maurya2025SelectLLM} adapts ensemble composition to the input while retaining a fixed cardinality. ThriftLLM~\cite{huang2025ThriftLLM} jointly adapts composition and cardinality under cost constraints, but addresses classification queries rather than open-ended response generation.

Other adaptive multi-model approaches make different inference decisions. EMS~\cite{liu2026EMS} expands the set of invoked models sequentially; LightRouter~\cite{zhang2025LightRouter} combines model selection with staged collaboration; and retrieval-based selection and response enhancement are investigated in~\cite{tang2026Scalable}. Query-specific agentic systems instead configure task plans, agents, or collaboration strategies~\cite{su2026difficulty,xiao2026RouterHGC,hamid2026optiq}. PISA-E targets pre-inference selection of a parallel ensemble for response generation. It scores candidate subsets using a strict-majority surrogate constructed from prompt-dependent individual capability estimates, jointly selecting composition and cardinality under an explicit cost--quality preference without training ensemble-specific predictors.

\section{Conclusion}
\label{sec:conclusion}


We introduced \router, a prompt-adaptive, cost-aware single-model router, and \routerM, its extension to ensemble routing. Both methods use prompt-dependent capability estimates, which \routerM reuses to jointly select ensemble composition and cardinality without ensemble-specific training. Across four heterogeneous benchmarks, \router achieves the highest macro-average hypervolume among the evaluated single-model routers under both ID and OOD conditions. \routerM achieves the highest macro-average task performance among the evaluated ensemble approaches, exceeding the strongest competitors by 2.09 and 2.87 percentage points, respectively, when prioritizing response quality. At selected lower-cost operating points, \routerM also improves macro-average task performance while reducing average inference cost by approximately 36\% under ID and 25\% under OOD conditions. Component analyses and ablations support the contributions of capability-aware response resolution, prompt-dependent ranking, and adaptive ensemble cardinality. Together, these results demonstrate that a shared capability-estimation framework supports effective single-model routing and selective ensembling, with cost–quality preferences adjustable at inference time without retraining.
For Web-based services such as search, recommendation, and intelligent agents, this flexibility offers a way to adapt inference expenditure to heterogeneous user requests and changing quality–cost preferences. 

\bibliographystyle{ACM-Reference-Format}
\bibliography{references}  

\appendix

\section{Additional Experimental Details}
\label{app:experimental-setup}

\paragraph{Benchmarks and evaluation scores.}
Table~\ref{tab:benchmark-details} summarizes the evaluated benchmarks. We use the original evaluation scores provided by each benchmark, which are all in the range $[0,1]$. FusionBench uses binary correctness, SPROUT employs LLM-judge scores, while RouterBench and Track-2 include task-dependent evaluation scores. Task performance is computed by averaging scores within each task group and then taking the macro-average across groups, consistently for all methods.
For capability estimation, \router{} derives binary success labels from the training scores using $y_{im}=\mathds{1}[s_{im}\geq 0.5]$ and trains model-specific predictors on these labels.

\begin{table}[t]
\centering\scriptsize
\caption{Number of prompts and candidate LLMs in each evaluation benchmark.}
\label{tab:benchmark-details}
\begin{tabular}{@{}lrr@{}}
\toprule
Benchmark & Prompts & Candidate models \\
\midrule
LLMRouterBench Track-2 & 12,446 & 13 \\
RouterBench & 34,684 & 11 \\
FusionBench & 6,398 & 40 \\
SPROUT & 44,241 & 13 \\
\bottomrule
\end{tabular}
\end{table}

\paragraph{Data partitions.}
For OOD evaluation, we construct group level partitions independently of the original benchmark splits. Entire task groups are assigned either to ID, which supplies training and validation data, or to OOD, which is reserved exclusively for testing. All available rows of held-out groups are used for OOD evaluation, irrespective of their original partition. The group assignments are selected without considering model outcomes, targeting approximately 20\% of the available prompts for OOD evaluation, subject to a 30\% upper bound. We require at least three OOD groups spanning three task categories and domains, and at least three retained ID groups. Among valid partitions, we select those closest to the target proportion. Table~\ref{tab:ood-partitions} reports the resulting assignments.

\begin{table}[t]
\centering
\scriptsize
\caption{Task-group partitions used for OOD evaluation. The final column reports the number and percentage of held-out prompts.}
\label{tab:ood-partitions}
\setlength{\tabcolsep}{2.5pt}
\renewcommand{\arraystretch}{1.08}
\begin{tabularx}{\columnwidth}{@{}l>{\raggedright\arraybackslash}X>{\raggedright\arraybackslash}X@{}}
\toprule
Benchmark & ID groups & OOD groups (rows; share) \\
\midrule
Track-2 & \texttt{aime}, \texttt{livecodebench}, \texttt{swe-bench}, \texttt{mmlupro}, \texttt{simpleqa}, \texttt{arenahard}, \texttt{tau2} & \texttt{livemathbench}, \texttt{gpqa}, \texttt{hle} (2,477; 19.9\%) \\
RouterBench & \texttt{mmlu}, \texttt{hellaswag}, \texttt{arc} & \texttt{winogrande}, \texttt{mbpp}, \texttt{grade} (9,144; 26.4\%) \\
FusionBench & \texttt{arc\_challenge}, \texttt{boolq}, \texttt{commonsense\_qa}, \texttt{gsm8k}, \texttt{hellaswag}, \texttt{human\_eval}, \texttt{mmlu}, \texttt{natural\_qa}, \texttt{openbook\_qa}, \texttt{squad}, \texttt{trivia\_qa} & \texttt{gpqa}, \texttt{math}, \texttt{mbpp} (1,278; 20.0\%) \\
SPROUT & \texttt{ragbench}, \texttt{openhermes}, \texttt{MMLU-Pro} & \texttt{MATH}, \texttt{MuSR}, \texttt{GPQA} (11,173; 25.3\%) \\
\bottomrule
\end{tabularx}
\end{table}

\paragraph{Experimental repetitions.}
The five random seeds are 11, 22, 33, 44, and 55. Thus, reported standard deviations measure run-to-run variability under fixed data partitions.

\paragraph{Baselines.}

DFPE follows its clustering and representative selection procedure, using global statistics fitted on ID training data when evaluating held-out OOD task groups. LLM-TOPLA constructs and prunes model teams following its original procedure. SelectLLM uses WeightedMaxConf, its best-performing configuration, with a RoBERTa-base multilabel classifier and inverse-square-root confidence weighted voting, following the original article. For PairRanker and MBR, fixed-size ensembles are formed from the top-$k$ models ranked by macro-average training performance.

\paragraph{Implementation Details.}
For \router{}, the maximum number of clusters and the soft-assignment parameter $\beta$ are selected using validation data. Cost--performance operating points are obtained by sweeping $\alpha$ from 0.10 to 1.00 in increments of 0.05.

\paragraph{Ensemble resolution.}
We consider several alternatives to the self-weighted BERTScore resolver introduced in Section~\ref{sec:methodology}. Given an ensemble $E$ and its generated responses $\{y_i\}_{i\in E}$, standard minimum Bayes risk (MBR) decoding with BERTScore~\cite{Zhang*2020BERTScore} selects the response maximizing its aggregate similarity to the ensemble outputs:
\begin{equation}
S_i^{\mathrm{MBR}}=\sum_{j\in E}\operatorname{BERTScore}(y_i,y_j).
\end{equation}
The other-weighted variant incorporates the prompt-dependent capability estimates produced by \router{}:
\begin{equation}
S_i^{\mathrm{other}}=\sum_{j\in E}w_j\operatorname{BERTScore}(y_i,y_j),
\end{equation}
where $w_j$ denotes the normalized capability estimate of the model generating response $y_j$. Unlike self-weighted BERTScore, other-weighted BERTScore does not additionally weight each candidate by its own estimated capability.

\paragraph{Inference costs.}
Routing decisions use training-set average model costs, normalized as described in Section~\ref{sec:methodology}. Reported inference costs instead use the benchmark-recorded costs of individual responses, summed over all selected models and averaged across test queries. Models with missing responses still contribute their recorded inference costs. Local routing and response-resolution overhead are excluded.

\paragraph{Hypervolume metric.}
We assess routing cost--performance trade-offs using normalized dominated hypervolume (HV)~\cite{zitzler1998HV}. Let $\mathcal{F}=\{(c_i,p_i)\}_{i=1}^{n}$ denote the non-dominated operating points ordered by increasing cost, with reference point $r=(c_{\mathrm{ref}},p_{\mathrm{ref}})$. The dominated area is
\begin{equation}
\mathrm{HV}_{\mathrm{raw}}=\sum_{i=1}^{n}(c_{i+1}-c_i)(p_i-p_{\mathrm{ref}}), \qquad c_{n+1}=c_{\mathrm{ref}},
\end{equation}
where only points within the reference bounds contribute. We normalize by
\begin{equation}
\mathrm{HV}_{\mathrm{norm}}=\frac{\mathrm{HV}_{\mathrm{raw}}}{c_{\mathrm{ref}}(1-p_{\mathrm{ref}})}.
\end{equation}
The performance reference is the macro-average expected score of a uniformly random single-model router. The cost reference is defined as a percentile of the per-prompt full-model-pool inference cost distribution, computed separately for each benchmark and evaluation partition. We use the 40th percentile for single-model routing comparisons and the 70th percentile for ensemble-routing comparisons. The higher reference for ensemble routing accommodates the larger inference costs associated with querying multiple models. Within each comparison, the same reference point is used for all methods. Both reference coordinates remain fixed across methods, and HV values are multiplied by 100 when reported.

\section{Gaussian-Copula Model for Dependent Ensemble Success}
\label{app:gaussian-copula}

The Poisson-binomial surrogate assumes conditional independence among model correctness events. To investigate whether modeling their dependencies improves ensemble selection, we consider a Gaussian-copula alternative.

For an ensemble $E$ of size $k$, let $\hat{p}_m(x)$ denote the capability estimate provided by \router{} for model $m$. We introduce latent variables $\mathbf{G}_E\sim\mathcal{N}(\mathbf{0},R_E)$, where $R_E$ is a correlation matrix, and define
\begin{equation}
B_m(x)=\mathds{1}\left[G_m\leq\Phi^{-1}(\hat{p}_m(x))\right],
\end{equation}
where $\Phi$ denotes the standard normal CDF. This construction preserves the marginal probabilities $\Pr(B_m(x)=1)=\hat{p}_m(x)$ while introducing dependencies through $R_E$. The ensemble-success surrogate becomes
\begin{equation}
\hat{p}^{\mathrm{GC}}_E(x)=\Pr\left(\sum_{m\in E}B_m(x)\geq\left\lfloor\frac{k}{2}\right\rfloor+1\right).
\label{eq:gc-majority}
\end{equation}
When $R_E=I$, the formulation reduces to the Poisson-binomial surrogate.

\paragraph{Dependence estimation.}

We estimate a prompt-independent correlation matrix $R$ from training data using pairwise maximum likelihood, following the composite-likelihood approach to multivariate probit estimation~\cite{ashford1970Probit}. 
Let $y_{im}=\mathds{1}[s_{im}\geq 0.5]$ denote the observed binary correctness of model $m$ on training prompt $i$, using the same threshold as \router{}, and let $\tilde{p}_{im}$ denote its out-of-fold capability estimate. For each model pair $(m,n)$, the probability of joint correctness under latent correlation $\rho$ is
\begin{equation}
\pi_{11}^{(i)}(\rho)=\Phi_2\left(\Phi^{-1}(\tilde{p}_{im}),\Phi^{-1}(\tilde{p}_{in});\rho\right),
\end{equation}
where $\Phi_2$ denotes the bivariate standard-normal CDF. The remaining outcome probabilities follow from the marginals: $\pi_{10}^{(i)}=\tilde{p}_{im}-\pi_{11}^{(i)}$, $\pi_{01}^{(i)}=\tilde{p}_{in}-\pi_{11}^{(i)}$, and $\pi_{00}^{(i)}=1-\tilde{p}_{im}-\tilde{p}_{in}+\pi_{11}^{(i)}$. We estimate each correlation as
\begin{equation}
\hat{\rho}_{mn}=\arg\max_{\rho\in(-1,1)}\sum_i\log\pi_{y_{im}y_{in}}^{(i)}(\rho).
\end{equation}
The pairwise estimates are assembled into $R$, with unit diagonal, and projected onto the nearest valid correlation matrix to ensure positive semidefiniteness.

\paragraph{Inference.}
For each candidate ensemble, we extract the corresponding submatrix $R_E$ and approximate Eq.~\eqref{eq:gc-majority} using 2,048 Monte Carlo samples from $\mathcal{N}(\mathbf{0},R_E)$. The estimated success probability is the fraction of samples satisfying the strict-majority condition. This estimate replaces the Poisson-binomial surrogate in the \routerM{} utility, leaving candidate construction, cost normalization, and ensemble selection unchanged.

\section{\router Additional Experiments}




\subsection{Hypervolume Reference-Point Sensitivity}
\label{app:router-hv-sensitivity}

\begin{figure}[th]
\centering
\includegraphics[width=\linewidth]{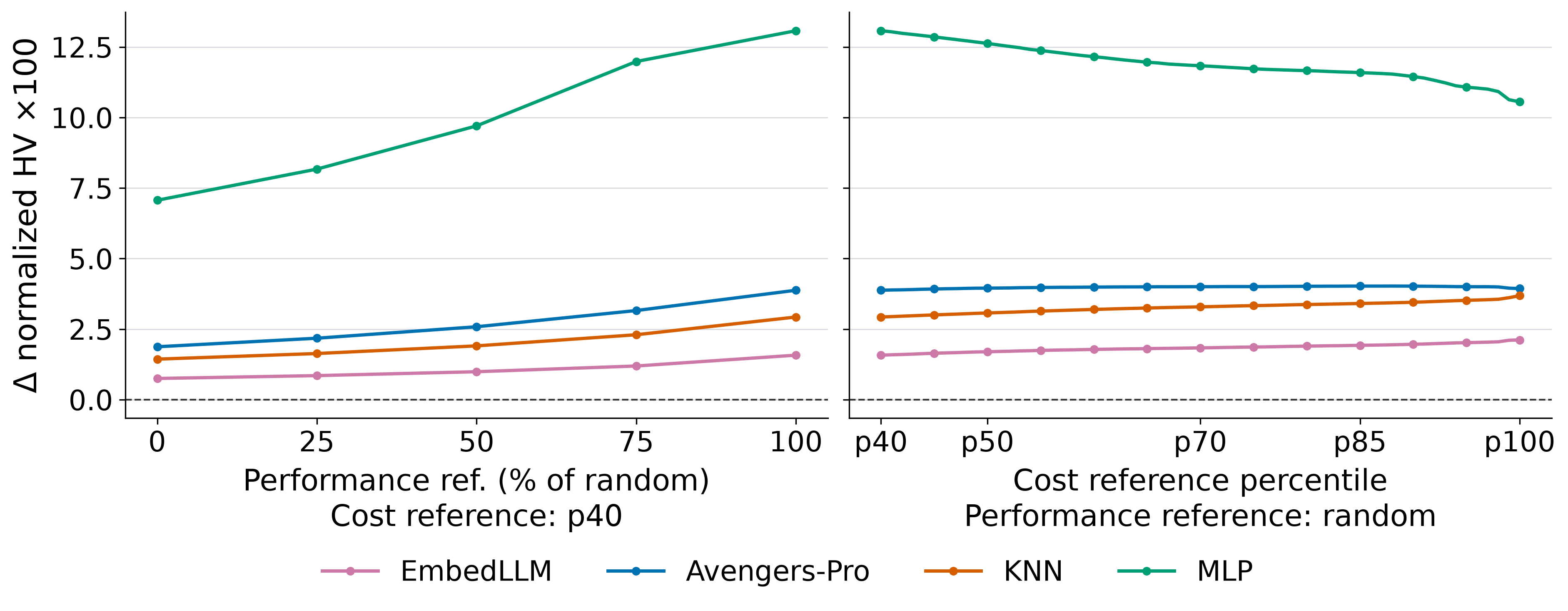}
\Description{}
\caption{Sensitivity of \router{}'s macro-average HV advantage over competing routers to the performance reference (left) and full-pool cost reference (right). Positive values favor \router{}.}
\label{fig:router-hv-sensitivity}
\end{figure}

To assess the sensitivity of our results to the hypervolume reference point, we vary both its cost and performance coordinates. As described in Appendix~\ref{app:experimental-setup}, the main experiments use fixed reference costs corresponding to the 40th and 70th percentiles of the full-pool cost distribution for single-model and ensemble routing, respectively. These reference points accommodate the operating points of the compared methods, while the reference performance corresponds to a uniformly random router. Although these choices provide a principled basis for comparison, HV values depend on the selected reference point. We therefore examine whether the aggregate advantage of \router{} persists under alternative reference points. Figure~\ref{fig:router-hv-sensitivity} shows that \router{} maintains a positive macro-average HV advantage over all competing single-model routers across the evaluated configurations. While the magnitude of this advantage varies with the reference coordinates, the relative ranking remains unchanged. These results indicate that the superior aggregate cost--performance trade-offs of \router{} are not contingent on the specific reference point adopted in the main evaluation.

\section{\routerM Additional Experiments}

\subsection{Hypervolume Reference-Point Sensitivity}
\label{app:fixed-k-hv-sensitivity}

\begin{figure}[t]
\centering
\includegraphics[width=\linewidth]{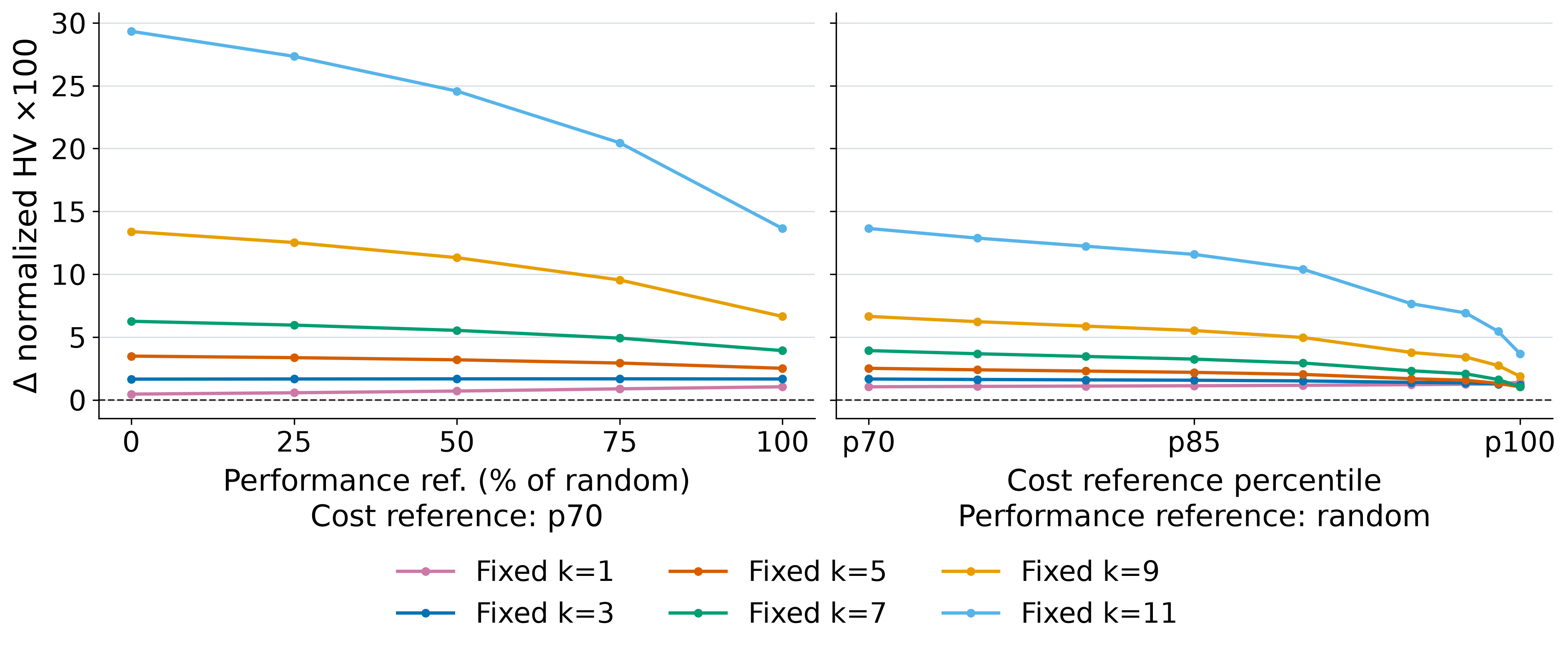}
    \Description{}
\caption{Sensitivity of \routerM{}'s macro-average HV advantage over fixed-cardinality variants to the performance reference (left) and full-pool cost reference (right). Positive values favor \routerM{}.}
\label{fig:ensemble-hv-sensitivity}
\end{figure}

We further examine whether the advantage of adaptive ensemble cardinality depends on the HV reference point selected for the main evaluation. Figure~\ref{fig:ensemble-hv-sensitivity} reports the macro-average HV differences between \routerM{} and each fixed-cardinality variant when varying the performance and cost references independently. Across all evaluated configurations, \routerM{} maintains a positive HV advantage over every fixed-cardinality alternative. Although the magnitude of these differences varies with the reference coordinates, the aggregate ranking remains unchanged.  The consistent advantage across reference configurations shows that the aggregate benefits of adaptive ensemble cardinality persist under different HV reference points.

\subsection{Sensitivity to Maximum Ensemble Cardinality}
\label{app:kmax-sensitivity}

\begin{table}[t]
\centering\scriptsize
\caption{Normalized HV ($\times 100$) of \routerM{} for different maximum ensemble cardinalities under ID and OOD conditions, using the p70 cost reference. Parentheses report sample standard deviations across five seeds.}
\label{tab:kmax-sensitivity}
\begin{tabularx}{\linewidth}{@{}l*{6}{>{\centering\arraybackslash}X}@{}}
\toprule
Scenario & $k_{\max}=1$ & $k_{\max}=3$ & $k_{\max}=5$ & $k_{\max}=7$ & $k_{\max}=9$ & $k_{\max}=11$ \\
\midrule
\rowcolor{black!8}\multicolumn{7}{l}{\textbf{ID}} \\
Track-2 & 27.89{\tiny\textcolor{black!45}{ (0.49)}} & 28.26{\tiny\textcolor{black!45}{ (0.64)}} & \ensuremath{\pmb{\mathbf{28.60}}}{\tiny\textcolor{black!45}{ (0.60)}} & \underline{28.58{\tiny\textcolor{black!45}{ (0.54)}}} & 28.51{\tiny\textcolor{black!45}{ (0.58)}} & 28.57{\tiny\textcolor{black!45}{ (0.59)}} \\
RouterBench & \ensuremath{\pmb{\mathbf{46.99}}}{\tiny\textcolor{black!45}{ (0.25)}} & 46.38{\tiny\textcolor{black!45}{ (0.26)}} & 46.25{\tiny\textcolor{black!45}{ (0.33)}} & 46.31{\tiny\textcolor{black!45}{ (0.29)}} & \underline{46.38{\tiny\textcolor{black!45}{ (0.28)}}} & 46.37{\tiny\textcolor{black!45}{ (0.31)}} \\
FusionBench & 42.74{\tiny\textcolor{black!45}{ (0.98)}} & 44.40{\tiny\textcolor{black!45}{ (0.69)}} & 44.68{\tiny\textcolor{black!45}{ (0.69)}} & 44.83{\tiny\textcolor{black!45}{ (0.72)}} & \ensuremath{\pmb{\mathbf{44.90}}}{\tiny\textcolor{black!45}{ (1.04)}} & \underline{44.89{\tiny\textcolor{black!45}{ (0.87)}}} \\
SPROUT & \ensuremath{\pmb{\mathbf{38.65}}}{\tiny\textcolor{black!45}{ (0.49)}} & 37.67{\tiny\textcolor{black!45}{ (0.82)}} & \underline{38.03{\tiny\textcolor{black!45}{ (1.21)}}} & 37.91{\tiny\textcolor{black!45}{ (1.05)}} & 37.60{\tiny\textcolor{black!45}{ (0.76)}} & 37.57{\tiny\textcolor{black!45}{ (0.67)}} \\
ID average & 39.07{\tiny\textcolor{black!45}{ (0.37)}} & 39.18{\tiny\textcolor{black!45}{ (0.48)}} & \underline{39.39{\tiny\textcolor{black!45}{ (0.50)}}} & \ensuremath{\pmb{\mathbf{39.41}}}{\tiny\textcolor{black!45}{ (0.48)}} & 39.35{\tiny\textcolor{black!45}{ (0.46)}} & 39.35{\tiny\textcolor{black!45}{ (0.44)}} \\
\midrule
\rowcolor{black!8}\multicolumn{7}{l}{\textbf{OOD}} \\
Track-2 & 20.54{\tiny\textcolor{black!45}{ (1.67)}} & 21.67{\tiny\textcolor{black!45}{ (0.80)}} & \ensuremath{\pmb{\mathbf{23.37}}}{\tiny\textcolor{black!45}{ (0.38)}} & \underline{22.51{\tiny\textcolor{black!45}{ (0.33)}}} & 22.42{\tiny\textcolor{black!45}{ (0.37)}} & 22.39{\tiny\textcolor{black!45}{ (0.34)}} \\
RouterBench & \ensuremath{\pmb{\mathbf{31.09}}}{\tiny\textcolor{black!45}{ (0.62)}} & 26.85{\tiny\textcolor{black!45}{ (0.84)}} & 24.77{\tiny\textcolor{black!45}{ (0.93)}} & \underline{27.04{\tiny\textcolor{black!45}{ (1.23)}}} & 26.83{\tiny\textcolor{black!45}{ (1.17)}} & 26.86{\tiny\textcolor{black!45}{ (1.29)}} \\
FusionBench & 23.03{\tiny\textcolor{black!45}{ (1.23)}} & 26.49{\tiny\textcolor{black!45}{ (0.38)}} & 27.81{\tiny\textcolor{black!45}{ (0.26)}} & 29.62{\tiny\textcolor{black!45}{ (0.33)}} & \underline{30.59{\tiny\textcolor{black!45}{ (0.09)}}} & \ensuremath{\pmb{\mathbf{30.96}}}{\tiny\textcolor{black!45}{ (0.22)}} \\
SPROUT & 23.32{\tiny\textcolor{black!45}{ (0.64)}} & 24.18{\tiny\textcolor{black!45}{ (0.31)}} & \ensuremath{\pmb{\mathbf{25.29}}}{\tiny\textcolor{black!45}{ (0.40)}} & \underline{25.11{\tiny\textcolor{black!45}{ (0.26)}}} & 25.04{\tiny\textcolor{black!45}{ (0.40)}} & 24.98{\tiny\textcolor{black!45}{ (0.44)}} \\
OOD average & 24.50{\tiny\textcolor{black!45}{ (0.31)}} & 24.80{\tiny\textcolor{black!45}{ (0.38)}} & 25.31{\tiny\textcolor{black!45}{ (0.25)}} & 26.07{\tiny\textcolor{black!45}{ (0.34)}} & \underline{26.22{\tiny\textcolor{black!45}{ (0.28)}}} & \ensuremath{\pmb{\mathbf{26.30}}}{\tiny\textcolor{black!45}{ (0.27)}} \\
\rowcolor{black!8}\multicolumn{7}{l}{\textbf{All scenarios}} \\
Full average & 31.78{\tiny\textcolor{black!45}{ (0.31)}} & 31.99{\tiny\textcolor{black!45}{ (0.32)}} & 32.35{\tiny\textcolor{black!45}{ (0.31)}} & 32.74{\tiny\textcolor{black!45}{ (0.32)}} & \underline{32.78{\tiny\textcolor{black!45}{ (0.25)}}} & \ensuremath{\pmb{\mathbf{32.82}}}{\tiny\textcolor{black!45}{ (0.25)}} \\
\bottomrule
\end{tabularx}
\end{table}

Table~\ref{tab:kmax-sensitivity} examines the effect of restricting the maximum ensemble cardinality $k_{\max}$. Under ID conditions, \routerM{} achieves a macro-average HV of 39.35 with $k_{\max}=11$, remaining competitive when the maximum cardinality is reduced, with HV reaching 39.41 at $k_{\max}=7$. Under OOD conditions, allowing larger ensembles provides more substantial benefits, although restricting $k_{\max}$ to seven still yields an HV of 26.07, compared with 26.30 when allowing up to eleven models. Thus, \routerM{} retains strong aggregate cost--performance trade-offs under moderate restrictions on ensemble cardinality, while the larger gains observed under OOD conditions highlight the benefits of access to a broader range of ensemble sizes.

\section{Greedy Selection vs. Exhaustive Surrogate Optimization}
\label{app:exact-vs-greedy}

\begin{figure}[t]
    \centering
    \includegraphics[width=\linewidth]{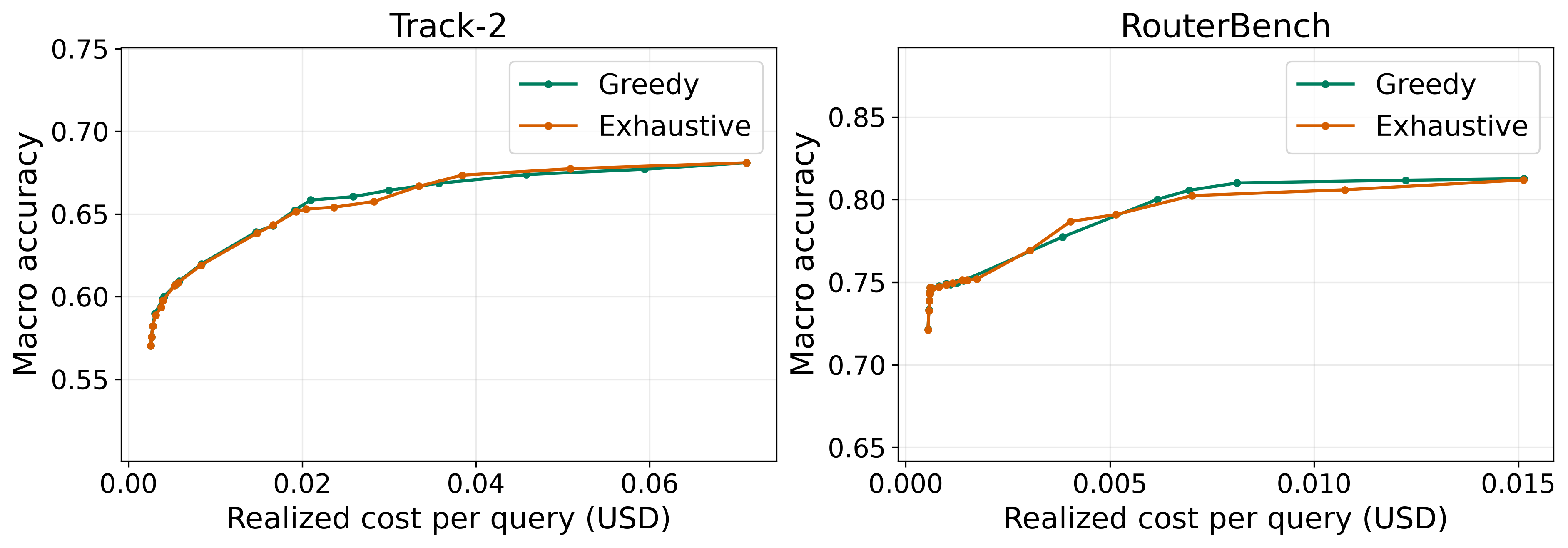}
    \Description{}
\caption{Cost--performance trade-offs of greedy candidate selection and exhaustive surrogate optimization on Track-2 and RouterBench, with $k_{\max}=7$.}
    \label{fig:exact-vs-greedy}
\end{figure}

We evaluate the greedy candidate-selection procedure by comparing it with exhaustive optimization of the ensemble-success surrogate. Both approaches use the same capability estimates, Poisson-binomial surrogate, and self-weighted BERTScore resolver, with $k_{\max}=7$. Exhaustive optimization evaluates all feasible model subsets, requiring the assessment of $\sum_{k=1}^{k_{\max}}\binom{M}{k}$ candidates for a pool of $M$ models. In contrast, greedy selection considers only $\lceil k_{\max}/2 \rceil$ odd-cardinality prefixes of the single-model utility ranking, substantially reducing the search space.

Figure~\ref{fig:exact-vs-greedy} compares the cost--performance curves on Track-2 and RouterBench. Greedy selection achieves trade-offs comparable to exhaustive optimization, with closely aligned curves on Track-2 and favorable performance across several intermediate-cost regions on RouterBench. These results indicate that greedy ranking prefixes provide a computationally efficient alternative while maintaining competitive cost--performance trade-offs in the evaluated settings.

\end{document}